\pdfoutput=1

\documentclass[11pt]{article}

\usepackage{acl}

\usepackage{times}
\usepackage{tabu}
\usepackage{latexsym}
\usepackage{stmaryrd}
\usepackage{amsmath}
\usepackage{multirow}
\usepackage{bbm}
\usepackage{graphicx}
\usepackage{amssymb}
\usepackage{booktabs}
\usepackage{adjustbox}
\usepackage{xspace}
\usepackage{algpseudocode}
\usepackage{algorithm}
\usepackage{graphicx}
\usepackage{caption}
\usepackage{subcaption}
\usepackage{tabularx}
\usepackage{colortbl}
\usepackage{xcolor}
\usepackage{tablefootnote}
\usepackage{enumitem}  
\usepackage{lipsum}
\usepackage{xspace}
\usepackage{bbding}
\usepackage{pifont}
\usepackage{arydshln}
\usepackage{array}
\usepackage{cleveref}
\usepackage{placeins}

\usepackage[T1]{fontenc}

\usepackage[utf8]{inputenc}

\usepackage{microtype}

\newcommand\instructgpt{\texttt{InstructGPT}~}
\newcommand\flant{\texttt{Flan-T5}~}
\newcommand\flantul{\texttt{Flan-UL2}~}
\newcommand\tkinstruct{\texttt{Tk-Instruct}~}
\newcommand\imlopt{\texttt{OPT-IML}~}
\newcommand\dolly{\texttt{Dolly-V2}~}
\newcommand\divp{\textit{DP}}
\newcommand\jigsaw{\textit{Jigsaw}}
\newcommand\diaz{\textit{Diaz}}
\newcommand\ghc{\textit{GHC}}
\newcommand\htwitter{\textit{H-Twitter}}
\newcommand\semeval{\textit{SE2016}}
\newcommand\gwsd{\textit{GWSD}}

\definecolor{PA}{rgb}{0.55, 0.71, 0.0}
\definecolor{All}{rgb}{0.91, 0.84, 0.42}
\definecolor{A}{rgb}{0.98, 0.38, 0.5}
\definecolor{R}{rgb}{0.64, 0.64, 0.82}
\definecolor{E}{rgb}{0.82, 0.62, 0.91}
\definecolor{G}{rgb}{0.03, 0.91, 0.87}

\definecolor{sociodemographics}{rgb}{0.0, 0.55, 0.55}
\definecolor{prompt}{rgb}{0.8, 0.36, 0.27}
\definecolor{dataset}{rgb}{0.72, 0.53, 0.04}
\newcommand{\colorline}[2]{{\color{#1}#2}}
\definecolor{tablegray}{rgb}{0.2,0.2,0.2}
\newcommand{\stdintable}[1] {~~\textcolor{tablegray}{\scriptsize{$\pm$#1}}}

\newcommand{\hs}[1]{#1}

\usepackage{todonotes}

\title{\emph{Sensitivity, Performance, Robustness:} \\Deconstructing the  Effect of Sociodemographic Prompting}

\author{Tilman Beck$^{1}$ \qquad Hendrik Schuff$^{1}$ \qquad Anne Lauscher$^{2}$ \qquad Iryna Gurevych$^{1}$\\
	$^1$Ubiquitous Knowledge Processing Lab (UKP Lab) \\
    Department of Computer Science and Hessian Center for AI (hessian.AI) \\
    Technical University of Darmstadt \\
        $^{2}$Data Science Group, University of Hamburg \\
	\texttt{\href{http://www.ukp.tu-darmstadt.de/}{www.ukp.tu-darmstadt.de}} \\
}

\begin{document}
\maketitle
\begin{abstract}
Annotators' sociodemographic backgrounds (i.e., the individual compositions of their \emph{gender}, \emph{age}, \emph{educational background}, etc.) have a strong impact on their decisions when working on subjective NLP tasks, such as toxic language detection. 
Often, heterogeneous backgrounds result in high disagreements.  
To model this variation, recent work has explored sociodemographic prompting, a technique, which steers the output of prompt-based models towards answers that humans with specific sociodemographic profiles would give. 
However, the available NLP literature disagrees on the efficacy of this technique — it remains unclear for which tasks and scenarios it can help, and the role of the individual factors in sociodemographic prompting is still unexplored.
We address this research gap by presenting the largest and most comprehensive study of sociodemographic prompting today. 
We analyze its influence on model sensitivity, performance and robustness across seven datasets and six instruction-tuned model families. 
We show that sociodemographic information affects model predictions and can be beneficial for improving zero-shot learning in subjective NLP tasks.
However, its outcomes largely vary for different model types, sizes, and datasets, and are subject to large variance with regards to prompt formulations.
Most importantly, our results show that sociodemographic prompting should be used with care for sensitive applications, such as toxicity annotation or when studying LLM alignment.\footnote{Code and data: \url{https://github.com/UKPLab/arxiv2023-sociodemographic-prompting}}

\end{abstract}

\section{Introduction}
How messages are perceived, is often not only dependent on their factual content, but also on the receiver's \emph{subjective interpretation}: for instance, two dataset annotators might have different equally valid opinions about what the ``correct'' offensiveness label for a particular tweet should be~\citep[e.g.,][\emph{inter alia}]{waseem-2016-racist, davani-etal-2023-hate}. 
As previously shown, this variation is, at least to some extent, tied to sociodemographic characteristics of the receivers, like their gender identity, age, and educational background~\citep[e.g.,][]{sap-etal-2022-annotators, biester-etal-2022-analyzing, pei-jurgens-2023-annotator}. 

\begin{figure}
    \centering
    \includegraphics[scale=0.6]{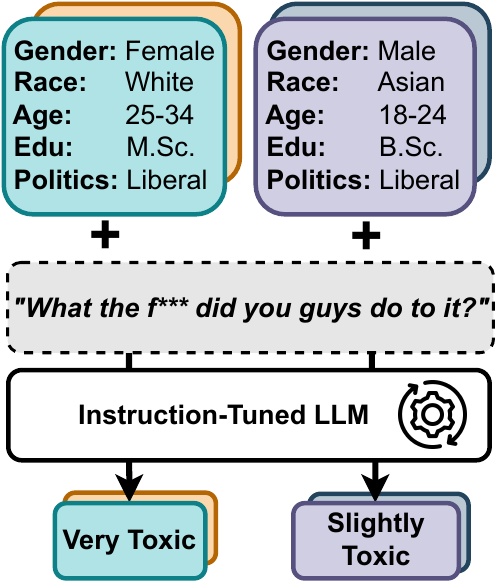}
    \caption{We instruct LLMs to make predictions for subjective NLP tasks from different perspectives using sociodemographic profiles. We show that, besides sociodemographics, outcomes are largely influenced by model choice or prompt formulation.}
    \label{fig:idea}
\end{figure}

Thus, NLP models need to account for a comprehensive set of sociodemographic factors to make more socially-aware predictions. 
Accordingly, modeling the effect of those factors on subjective tasks has emerged as an important research direction for NLP. 
As such, researchers have proposed new data collection paradigms -- cf. \emph{perspectivism}~\citep{rottger-etal-2022-two} -- and trained models for reflecting the decisions of particular sociodemographic groups~\citep{gupta_samesame2023, fleisig2023majority}. 
Most recently, researchers~\citep{deshpande2023toxicity, santurkar2023opinions, hwang2023aligning, cheng-etal-2023-marked} have explored \emph{sociodemographic prompting} of large language models (LLMs):  the idea is to enrich a particular input prompt with additional sociodemographic information (cf. Figure~\ref{fig:idea}). 
The models' output should then be aligned with the population described.
This technique has led to promising applications in dataset augmentation~\citep{hartvigsen-etal-2022-toxigen} and simulation of social computing platforms~\citep{socialsimulacra2022}.
Still, our knowledge on the effect of including sociodemographic profiles is scarce and the existing literature disagrees on its usefulness: for instance, \citet{argyle2023} showed that sociodemographic prompting can be used to simulate human populations -- a promise for more efficient sociological surveys. 
Other work, in turn, points to the danger of stereotypical bias reflected when prompting models with sociodemographic profiles~\citep{cheng-etal-2023-marked, deshpande2023toxicity}.

Our work makes an important step towards a better understanding of the influence of sociodemographic prompting on models' decisions.
We use subjective NLP tasks for evaluation as they have been shown to induce disagreement among annotators related to sociodemographic factors.
Our study analyzes the effects of sociodemographic prompting on model output (\emph{sensitivity}), zero-shot \emph{performance}, and \emph{robustness}. 
Concretely, we make the following contributions:
\begin{itemize}
    \item We present the largest and most comprehensive study on sociodemographic prompting to-date. Concretely, we test the effect of instructing 17 LLMs  (covering various model types, e.g., \texttt{InstructGPT}, \texttt{Flan-T5}, etc.) with sociodemographic profiles in a controlled setting which comprises seven datasets reflecting four different subjective NLP classification tasks (sentiment analysis, hatespeech detection, toxicity detection, and stance detection).
    \item We demonstrate (§\ref{ssec:model_sensitivity}) that sociodemographic prompting leads to surprisingly large amounts of prediction changes (up to 80\%), with large variance across model types and sizes.
    \item Our findings (§\ref{ssec:zeroshot}) indicate that sociodemographic prompting helps both to classify annotator-specific decisions and in zero-shot learning with performance improvements up to +8pp in accuracy.
    \item We show (§\ref{ssec:robustness}) that sociodemographic prompting is not robust, with large variance due to prompt formulation and model choice.
\end{itemize}

Overall, our results provide important insights for future research on sociodemographic prompting, and, in particular, when applying sociodemographic prompting in sensitive scenarios, for instance, in the context of sensitive data annotation or when studying LLM alignment.

\section{Related Work}

The sociodemographic background of annotators has been identified as an influential factor in text annotation for subjective NLP tasks~\citep{luo-etal-2020-detecting, sap-etal-2022-annotators, biester-etal-2022-analyzing, pei-jurgens-2023-annotator, santy2023nlpositionality}.
Consequently, researchers started to integrate such information into NLP models to enable more socially aware predictions~\citep{kumar2021designing, gordon2022, gupta_samesame2023, fleisig2023majority, Wan_Kim_Kang_2023}.

With the increasing performance of LLMs, researchers investigated to what extent they are influenced when prompted with sociodemographic information.
\citet{lee2023large} examine whether instruction-tuned LLMs accurately reflect or conform to human disagreements but limit their experiments to a single NLI dataset.
They conclude that models deviate from human annotators in terms of accuracy and disagreement level.
By analyzing disagreement for Q\&A, \citet{hwang2023aligning} find that users' opinions and their sociodemographic background are not mutual predictors.
For predicting individual opinions, they show that combining sociodemographic information and relevant past opinions performs best. 
Several works~\citep{durmus2023measuring, santurkar2023opinions, santy2023nlpositionality} analyze LLM's alignment with specific sociodemographic groups and show that model responses are biased towards responses by participants from Western countries.
Notably, \citet{santurkar2023opinions} observe that misalignment persists even after explicitly steering the LMs towards particular demographic groups.
\citet{argyle2023} suggest using GPT-3 as testbed before conducting large-scale population surveys. 
They propose \textit{algorithmic fidelity} to evaluate alignment with different human subpopulations and present it as a cost-efficient proxy for specific human sub-populations in social science research.
Finally, it has been shown~\citep{cheng-etal-2023-marked, deshpande2023toxicity, ungless-etal-2023-stereotypes, attanasio-etal-2023-tale} that prompting large models with sociodemographic information is prone to amplify existing stereotypical biases~\citep[cf.][]{blodgett-etal-2020-language, barikeri-etal-2021-redditbias}.

In contrast to studying LLM alignment, our work extends previous work showing that integrating sociodemographic information in a \textit{supervised} manner is helpful to improve annotator-specific predictions~\citep{gupta_samesame2023, fleisig2023majority}.
We use sociodemographic prompting because it allows to diversify model predictions without the need for privacy-concerning data collection of annotators' sociodemographic information.
Beyond our objective, our results offer important insights into influential factors for sociodemographic prompting in general.

\section{Sociodemographic Prompting}
\label{sec:sociodemographic_prompting}

Throughout this work, we \emph{prompt} a language model with (or without) \emph{sociodemographic information} for \emph{obtaining predictions} for the classification tasks we study. 
In the following, we discuss the main concepts our methodology relies on.

\begin{table*}[t!]
    \centering
    \resizebox{1.00\textwidth}{!}{%
    \begin{tabular}{p{0.2\textwidth}|p{0.8\textwidth}}
        \toprule
        \textbf{Attribute} & \textbf{Values (Percentage share)} \\
        \midrule
        Gender & male (52\%), female (47\%), nonbinary (<1\%) \\
        Race & White (77\%), Black or African American (13\%), Asian (6\%), Hispanic (3\%), Native Hawaiian or Pacific Islander (1\%), American Indian or Alaska Native (<1\%) \\
        Age & Under 18 (<1\%), 18 - 24 (11\%), 25 - 34 (40\%), 35 - 44 (25\%), 45 - 54 (13\%), 55 - 64 (8\%), 65 or older (3\%) \\
        Education & Less than high school degree (1\%), High school graduate (9\%), Some college but no degree (19\%), Associate degree in college (2-year) (11\%), Bachelors degree in college (4-year) (42\%), Masters degree (16\%), Professional degree (JD, MD) (2\%), Doctoral degree (1\%) \\
        Political Affiliation  & Liberal (43\%), Conservative (29\%), Independent (28\%) \\
        \bottomrule
    \end{tabular}
    }
    \caption{The sociodemographic attributes and their corresponding values we use in this study, based on the DP dataset by \citet{kumar2021designing}. Ordered ordinally by qualitative scale or by percentage share.}
    \label{tab:sd-attributes}
\end{table*}

\paragraph{Prompting.} Prompting refers to the act of providing an initial input or cue to a language model, guiding its subsequent output generation.
LLMs rely on these prompts to produce contextually relevant and coherent responses.

\paragraph{Sociodemographic Information.} Sociodemographic information encompasses data related to the social and demographic characteristics of individuals or groups. 
Providing sociodemographic cues via prompting has been shown to influence LLM's responses~\citep{durmus2023measuring, hwang2023aligning} through leveraging the sociodemographic knowledge encoded in the models~\citep{lauscher-etal-2022-socioprobe}. 
The promise is that the output will be tailored to a specific demographic or social group. 
The present analysis encompasses five distinct sociodemographic attributes, based on the datasets we employ: gender, race, age range, education level, and political affiliation, as detailed in Table~\ref{tab:sd-attributes}. 
In Figure~\ref{fig:sd-prompt-example}, we provide an example of a sociodemographic prompt.

\begin{figure}[!t]
\fbox{%
	\begin{minipage}{0.465\textwidth}
	\colorline{prompt}{Given a text, how would a person of gender} \colorline{sociodemographics}{'Female', race 'White', age '25 - 34', education level 'Master's degree' and political affiliation 'Liberal'} \colorline{prompt}{rate the degree of toxicity in the text. Possible values are} \colorline{dataset}{'not toxic', 'slightly toxic', 'moderately toxic', 'very toxic' or 'extremely toxic'}.
    
    \colorline{prompt}{Text}: \colorline{dataset}{'Well when you have a welfare state that propagates an underclass of unskilled parasites'}
    
    \colorline{prompt}{Toxicity}:
    \end{minipage}
    }
    \caption{Sociodemographically enriched prompt to predict the level of toxicity in a text. The different parts of the prompt are highlighted, i.e. \colorline{prompt}{instruction}, \colorline{sociodemographics}{sociodemographic properties} and \colorline{dataset}{dataset input}. Example drawn from the  dataset by \citet{kumar2021designing}.}
    \label{fig:sd-prompt-example}
\end{figure}

\paragraph{Obtaining Predictions.} Our strategy for answer generation differs between closed-source LLMs and open-source alternatives. 
For open-source models, we follow existing work~\citep{brown2020language, ye2023compositional} and prompt the model independently for each possible label by appending the potential answer to the prompt. 
Then, we evaluate the likelihood associated with each option and select the one with the highest likelihood. 
In scenarios requiring binary classification, we assign semantically coherent descriptors to each label, e.g.,  \emph{``Yes''} or \emph{``No''} in lieu of 0 or 1 for binary hate speech detection.
For closed-source models, we post-process the model output and map it to the predefined label space, to reduce the number of required API calls.
In the few cases where this approach fails, we assign it manually.

\section{Overall Experimental Setup}
\subsection{Tasks and Datasets}

\begin{table*}
    \centering
    \resizebox{1.00\textwidth}{!}{%
    \begin{tabular}{l|l|l|c}
        \toprule
        \textbf{Task} & \textbf{Dataset} & \textbf{Labels} & \textbf{IAA}  \\
       \midrule
       Toxicity & DP & not toxic (52\%), slightly toxic (19\%), moderately toxic (14\%), very toxic (9\%), extremely toxic (6\%) & 0.13 \\
        & Jigsaw & yes (67\%), no (33\%) & 0.46 \\
       Hatespeech & GHC & yes (87\%), no (13\%) & 0.25 \\ 
       & H-Twitter & neither (79\%), sexism (17\%), racism (3\%), both(1\%)  & 0.59 \\
       Stance & SE2016 & against (55\%), none (23\%), favor (22\%) & 0.58 \\
       & GWSD & agree (38\%), neutral (44\%), disagree (18\%) & 0.33 \\ 
       Sentiment & Diaz & very positive (9\%), somewhat positive (24\%), neutral (41\%), somewhat negative (21\%), very negative (5\%) & 0.11 \\ 
       \bottomrule
    \end{tabular}}
    \caption{The tasks and datasets (\textit{Diverse Perspectives (\divp{})}, \jigsaw{}, \semeval{}, \emph{Global Warming Stance Detection (\gwsd{})}, \emph{Gabe Hate Corpus (\ghc{})}, \emph{Twitter Hatespeech Corpus (\htwitter{})}, and \emph{Diaz}) we use along with their labels and inter-annotator agreement we obtain (IAA, Krippendorff's $\alpha$).}
    \label{tab:datasets}
\end{table*}

We select seven datasets for four subjective tasks (toxicity detection, stance detection, hatespeech detection, and sentiment classification) to study sociodemographic prompting across a large and diverse benchmark (cf. Table~\ref{tab:datasets},  Appendix~\ref{app:sec:datasets}).
Human annotations for those tasks have been shown to be influenced by sociodemographic factors.
Some datasets have been specifically proposed for tuning NLP systems, others have been published to analyze annotator disagreement, which explains the variability in IAA.

We have access to the original, un-aggregated annotations for each dataset. 
To analyze the effect of sociodemographic prompting we additionally require sociodemographic profiles. 
For two of the datasets (\divp{}, \emph{Diaz}) we have access to this information and adhere to the original sociodemographic details for prompting. 
For the remaining five datasets, we adopt the sociodemographic profiles of the annotators of the toxicity dataset \divp{} as a replacement, as done by \citet{Wan_Kim_Kang_2023}.
In particular, each example gets a set of five profiles, of which each is a composition of different sociodemographic attribute values.  

For all datasets, we removed instances with incomplete or unknown information (details in Appendix~\ref{app:sec:datasets}).
Due to the large number of experiments and varying dataset sizes, we randomly sample 1,000 instances from each dataset.
Our sampling strategy leads to a similar distribution across the different sociodemographic attributes, as we demonstrate in Appendix~\ref{app:sec:sampling}.
In the following, we describe the individual datasets.

\paragraph{Toxicity.} The task is to decide whether or to what degree (e.g., \emph{slightly toxic}) a text is toxic. 
We utilize \textit{Diverse Perspectives}~(DP) by \citet{kumar2021designing}, and \jigsaw{}~\citep{goyal2022toxicity}.

\divp{} comprises comments from various online forums (e.g., 4chan, Reddit).
These comments underwent annotation via Amazon Mechanical Turk, receiving five annotations per instance. 
For each annotator, the sociodemographic data was gathered.
The dataset did not come equipped with a definitive gold label. 
Thus, we use majority voting to determine the gold label.

\jigsaw{} encapsulates comments from news articles, originally collated by the \textit{Civil Comments} platform and subsequently annotated for toxicity indicators.  
The binary gold label for this dataset was derived by classifying comments as toxic if a majority of annotators identified them as such.

\paragraph{Stance.} Stance detection, pertains to discerning an author's viewpoint towards a specific topic~\citep{stance_survey2020, beck-etal-2021-investigating, lauscher-etal-2022-scientia}. 
As shown by \citet{balahur-etal-2010-sentiment} and \citet{luo-etal-2020-detecting}, annotators' decisions are influenced by their sociodemographic background.
We employ the SemEval 2016 Task 6 dataset \citep[SE2016;][]{StanceSemEval2016} and the \textit{Global Warming Stance Detection} (GWSD) dataset \citep{luo-etal-2020-detecting}.

\semeval{} encompasses 3,591 annotated Twitter posts that address a range of contentious subjects. 
The gold labels were ascertained using majority voting. 
Instances exhibiting less than 60\% consensus among annotators were excluded by the authors.

\gwsd{} consists of 2,050 annotated U.S. news articles and was curated to analyze the framing of opinions within the discourse on global warming. 
To determine the gold label for each article, the authors employed a model tailored to the distribution of annotations, which also factored in potential biases of the annotators.

\paragraph{Hatespeech.} Hatespeech detection is a task designed to tackle the increasing amount of hateful online communication.
We use the \emph{Gabe Hate Corpus} (GHC) by \citet{kennedy2022introducing} and the \emph{Twitter Hatespeech Corpus}~\citep[H-Twitter;][]{waseem-2016-racist}.
 
\ghc{} was sourced from the social network service gab.com and annotated in a multi-label fashion for \textit{Human Degradation}, \textit{Calls For Violence} and \textit{Vulgar/Offensive}.
The authors obtained gold labels using majority voting.
As we are comparing multi-class tasks, we binarized the annotations into hatespeech indicators (i.e., \textit{Yes} and \textit{No}).

\htwitter{} was annotated by CrowdFlower workers for \emph{sexism}, \emph{racism}, \emph{neither}, or \emph{both}. 
Expert annotators
contributed the gold labels. 

\paragraph{Sentiment.} 
The task is to decide upon the sentiment conveyed in the text. 
We use the dataset by \citet{diaz2018addressing}, which we call \emph{Diaz}, created for studying age-related bias in sentiment analysis. 

\subsection{Models}

We seek to \emph{instruct} models to mimic an annotator with a specific sociodemographic profile.
Thus, we resort to the most natural choice, instruction-tuned models. 
We aim to cover a broad collection of models, both from industrial and academic research and fine-tuned using different instruction-tuning datasets.
If possible, we chose model families with several model sizes published.
Concretely, we use GPT-3~\citep{brown2020language}, T5~\citep{raffel2020exploring}, OPT~\citep{zhang2022opt}, and Pythia~\citep{biderman2023pythia} model variants. 
We present a comprehensive overview of all models in Appendix~\ref{app:sec:modeldetails}.  

\paragraph{GPT-3.} We use \instructgpt~\citep{ouyang2022training} which was fine-tuned using reinforcement learning from human feedback (RLHF). 

\paragraph{T5.} We further use  \flant~\citep{chung2022scaling}, \flantul~\citep{tay2023ul} and \tkinstruct~\citep{wang-etal-2022-super}.
\flant was trained over a collection of 1,836 finetuning tasks.
\flantul uses the same instruction-tuning procedure but is built on top of a language model which was trained following the Unifying Language Learning Paradigm (UL2) pretraining framework.
\flant (\tkinstruct) were trained using large benchmark of 1,836 (1,616) NLP tasks and their natural language instructions.
\paragraph{OPT.} We employ \imlopt~\citep{iyer2023optiml} which was fine-tuned using an aggregation of eight instruction-tuning datasets comprising 1,991 tasks.

\paragraph{Pythia.} 
We use \dolly which is fine-tuned on a 15K instruction corpus\footnote{\url{https://tinyurl.com/databricks-dolly}} covering seven task categories.

\subsection{Evaluation}

For subjective NLP tasks~\citep{ovesdotter-alm-2011-subjective}, comparing aggregated annotations with model predictions provides only a limited view on the performance as label aggregation obscures any disagreement in the data~\citep{prabhakaran-etal-2021-releasing, basile-etal-2021-need}.
Thus, we follow \citet{uma2021learning} and evaluate our results using both hard-label evaluation (accuracy, macro-averaged F1) and soft-label evaluation (cross-entropy, Jensen-Shannon divergence).
In case of hard-label evaluation, we aggregate all predictions obtained via sociodemographic prompting using majority voting.
To test for statistical significance of our results, we use generalized linear mixed models (GLMMs) to account for potential confounders and statistical dependencies in our data by jointly modeling numerous main effects (e.g., the impact of model family) and interaction effects (e.g., the joint impact of model family and prompting method).
We report further details about our experiments (§\ref{app:sec:experimental_details}) and the statistical analysis using GLMMs (§\ref{sec:glmm_analyses}) in the Appendix.

\begin{figure}
    \centering
    \includegraphics[width=.5\textwidth]{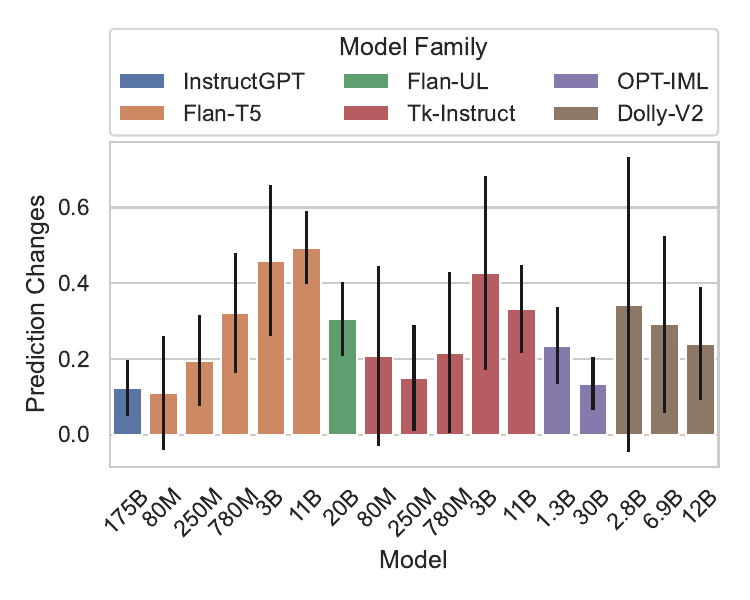}
    \caption{Mean percentage of prediction changes across all datasets when comparing outputs of zero-shot prompting with and without sociodemographic information. The x-axis denotes the model size and the color indicates the model family.}
    \label{fig:label_changes}
\end{figure}

\section{Sensitivity}
We investigate how \emph{sensitive} the predictions are, i.e., to what extent LLMs' predictions change when instructed to answer from viewpoints characterised by particular sociodemographic backgrounds.

\label{ssec:model_sensitivity}
\subsection{Detailed Setup}  
We aggregate all predictions from prompting with different profiles using majority voting.
Then, we compare how often the aggregated label differs from the one predicted without any sociodemographic information.

\subsection{Results}
\paragraph{Prompting using sociodemographic profiles leads to prediction changes.}
\label{para:label_changes}

In Figure~\ref{fig:label_changes}, we depict the amount of label prediction changes (in percent) when including sociodemographic information, averaged across all datasets. 
Several trends can be observed; the degree of prediction changes is both dependent on the model and dataset.
Notably, instruction-tuned models based on T5 (\texttt{Flan-T5}, \texttt{Tk-Instruct}) or Pythia (\texttt{Dolly-V2}) are on average more affected by sociodemographic prompting than \instructgpt or variants of \imlopt{}.

Interestingly, we find that prediction changes are statistically significantly affected by the length of the text instance, i.e. shorter texts are associated with an increased number of prediction changes.

Looking at individual datasets (Figure~\ref{app:fig:label_changes}) we observe that models are more affected by data from \divp{} and \diaz{}, with extreme cases where more than 80\% of the predictions change when using \dolly (2.8B).
In contrast, hatespeech datasets show less pronounced label shifts.

\paragraph{Model choice and text ambiguity are influential factors}
To better understand the reasons for prediction changes, we aim at analyzing which textual properties lead to prediction changes when prompting with different sociodemographic profiles.
We are interested in model-agnostic reasons for sensitivity of sociodemographic prompting.
Thus, to draw valid conclusions, we filtered those instances which led to prediction changes across all models tested.
Surprisingly, none of the changes are consistent across all models.
This indicates that the model choice has a large influence on the prediction outcome, an observation which we inspect closer in §\ref{ssec:performance_results}.

We further suspect that ambiguity in the text (i.e. disagreement among annotators) might be a reason for prediction changes.
We compute the correlation of the disagreement observed in the original annotations and among the results using different sociodemographic profiles.
We use the variance to mean ratio as our proxy score for the level of disagreement per instance.
We observe a weak Kendall's $\tau$ \cite{kendall1938new} correlation which is statistically significant ($\tau=.07$, $p<.001$).

\paragraph{Combinations of sociodemographic attributes are more influential than individual attributes in isolation.}

We investigate the differences between using several sociodemographic attributes and using only one attribute at a time.
In general, the combination is most influential, i.e., in 63\% of the experiments across models and datasets it leads to most prediction changes.
However, there are some dataset-specific effects across different model families where individual attributes have a stronger impact, e.g. \textit{race} for the \ghc{} corpus (detailed results in Appendix~\ref{app:sec:label_changes_per_attribute}).

\section{Performance}
\label{ssec:zeroshot}

\begin{table}
    \centering
    \resizebox{.49\textwidth}{!}{%
    \begin{tabular}{l|cc|cc}
    \toprule
    & \multicolumn{2}{c}{Toxicity - DP} & \multicolumn{2}{c}{Sentiment - Diaz} \\
         Model & Acc & F1 & Acc & F1 \\
    \midrule
    Random & .19 & .17 & .20 & .17 \\
    Majority & .06 & .02 & .09 & .03 \\
    \midrule
    InstructGPT(175B) & .43 & .26 & .34 & .26 \\
    InstructGPT(175B)+SD & \textbf{.44} & .26 & \textbf{.37} & \textbf{.31} \\
    \midrule
    OPT-IML(30B) & .42 & .18 & .28 & .26 \\
    OPT-IML(30B)+SD & \textbf{.45} & .18 & \textbf{.32} & \textbf{.27} \\
    \bottomrule
    \end{tabular}
    }
    \caption{Zero-shot performance when predicting annotator-specific annotations using the original sociodemographic profile. We compare prompting with (SD) and without sociodemographic information and report macro-averaged F1 and Accuracy (Acc). Bold scores highlight the better performance when comparing the same model.}
    \label{tab:zeroshot_original_annotations}
\end{table}

We analyze the impact of sociodemographic prompting on models' \emph{zero-shot performances}.

\subsection{Detailed Setup} 
First, we evaluate to what extent sociodemographic prompting predicts an annotator-specific annotation with the same sociodemographic profile as provided in the prompt. 
We study this setup for the datasets where this information is provided (\divp{}, \diaz{}). 
Second, we evaluate the performance of sociodemographic prompting by comparing it to the original set of annotations for each dataset.
In the following, we present the results for the two overall best-performing models, \instructgpt and \imlopt, and provide the complete results in Appendix~\ref{app:sec:zeroshot_performance}.
Note, that statistical analyses were performed on experimental results from all models.

\subsection{Results}
\label{ssec:performance_results}
\paragraph{Adding sociodemographic information helps reproducing individual annotator decisions.}

\begin{table*}
    \centering
    \resizebox{1.00\textwidth}{!}{%
    \begin{tabular}{lcccccccccccccccccccccccccccc}
    \toprule
          & \multicolumn{4}{c}{Toxicity} & \multicolumn{4}{c}{Hatespeech} & \multicolumn{4}{c}{Stance Detection} & \multicolumn{2}{c}{Sentiment}  \\
         & \multicolumn{2}{c}{DP} & \multicolumn{2}{c}{Jigsaw} & \multicolumn{2}{c}{GHC} & \multicolumn{2}{c}{H-Twitter} & \multicolumn{2}{c}{SE2016} & \multicolumn{2}{c}{GWSD} & \multicolumn{2}{c}{Diaz} \\
         Model & Acc & F1 & Acc & F1 & Acc & F1 & Acc &F1 & Acc &F1 & Acc &F1 & Acc & F1 \\
         \midrule
    InstructGPT(175B)&.48&.27&.76&.59&.89&\textbf{.75}&\textbf{.87}&\textbf{.57}&\textbf{.53}&.51&\textbf{.72}&\textbf{.69}&.36&.33\\
    InstructGPT(175B) +SD&\textbf{.51}&\textbf{.28}&\textbf{.79}&\textbf{.60}&.89&.74&.86&.53&.52&\textbf{.52}&.69&.68&\textbf{.39}&.33\\
    \midrule
    OPT-IML(30B)&.50&.19&.58&.49&.80&.69&.84&.54&\textbf{.67}&\textbf{.53}&\textbf{.57}&\textbf{.50}&.27&.24\\
    OPT-IML(30B) SD&\textbf{.55}&\textbf{.20}&\textbf{.64}&\textbf{.53}&\textbf{.85}&\textbf{.74}&\textbf{.86}&\textbf{.57}&.65&.51&.52&.39&\textbf{.35}&\textbf{.29}\\
    \midrule
     & CE & JSD & CE & JSD & CE & JSD & CE & JSD & CE & JSD & CE & JSD & CE & JSD \\
    \midrule
    InstructGPT(175B)&1.42&.32&.55&.15&.43&.07&\textbf{.88}&.08&\textbf{.98}&.27&\textbf{.86}&.18&1.50&.37\\
InstructGPT(175B)+SD&\textbf{1.40}&\textbf{.29}&\textbf{.51}&\textbf{.12}&\textbf{.42}&\textbf{.06}&.90&.08&.99&\textbf{.25}&.89&\textbf{.15}&\textbf{1.48}&\textbf{.33}\\
    \midrule
OPT-IML(30B)&1.43&.33&.71&.26&.52&.13&.90&.10&\textbf{.90}&.22&\textbf{.95}&.23&1.57&.42\\
OPT-IML(30B)+SD&\textbf{1.40}&\textbf{.29}&\textbf{.66}&\textbf{.20}&\textbf{.48}&\textbf{.09}&\textbf{.89}&\textbf{.07}&.91&\textbf{.21}&.99&.23&\textbf{1.52}&\textbf{.32}\\
    
\bottomrule
    \end{tabular}
    }
    \caption{Comparison of zero-shot prompting performance using hard-label evaluation (Accuracy, F1) and soft-label evaluation (Cross-Entropy as CE, Jensen-Shannon Divergence as JSD), with (SD) and without sociodemographic information. Bold scores highlight the better performance when comparing the same model.}
    \label{tab:zeroshot_performance}
\end{table*}

Table~\ref{tab:zeroshot_original_annotations} demonstrates a positive trend when providing the original sociodemographic profiles.
This holds true for larger models (>11B). 
Most models from other families (\tkinstruct or \dolly) do not outperform random prediction, independent of the prompting method (cf. Table~\ref{app:tab:zeroshot_original_annotations}).
Still, predicting annotator-specific decisions is challenging with more than half of the instances (Accuracy < .5) being incorrectly classified.
This is partially due to the datasets' label imbalance, as indicated by the relatively low F1 scores.
These results confirm to some extent earlier work stating that sociodemographic information may not provide enough information to explain individual annotation behavior~\citep{crowdworksheets_diaz2022, orlikowski-etal-2023-ecological}.
Interestingly, our statistical analyses show that sociodemographic prompting has a significant interaction effect with input text,  i.e. it is more effective for longer input texts.

\paragraph{Sociodemographic prompting can improve zero-shot performance.}

Table~\ref{tab:zeroshot_performance} presents the hard-label and soft-label evaluation results.
There is a statistically significant interaction effect of model family and prompting method, identifying \instructgpt as the model which benefits most from sociodemographic prompting and \flant least.
Interestingly, for toxicity detection and sentiment classification, the models benefit from sociodemographic prompting, whereas for stance detection they perform better without such information. 
We observe a slight trend that datasets for which improvements are observed share low IAA across the original annotations (see Krippendorff's $\alpha$ in Table~\ref{tab:datasets}).
Most notably, using the sociodemographic profiles from the \divp{} dataset can also improve performance for other datasets such as \jigsaw{}, \ghc{} and \gwsd{}.

When comparing both evaluation setups, the positive effect of sociodemographic prompting is more pronounced for soft-label evaluation. 
This indicates that, overall, the predictions are more aligned to the original annotations. 
The results for the other model sizes are provided in  Appendix~\ref{app:sec:zeroshot_performance}.
We observe that multiple model configurations exhibit weak performance for both setups in general, often without any increasing trend for larger models from the same model family.

\begin{figure}
    \centering
    \includegraphics[width=.5\textwidth]{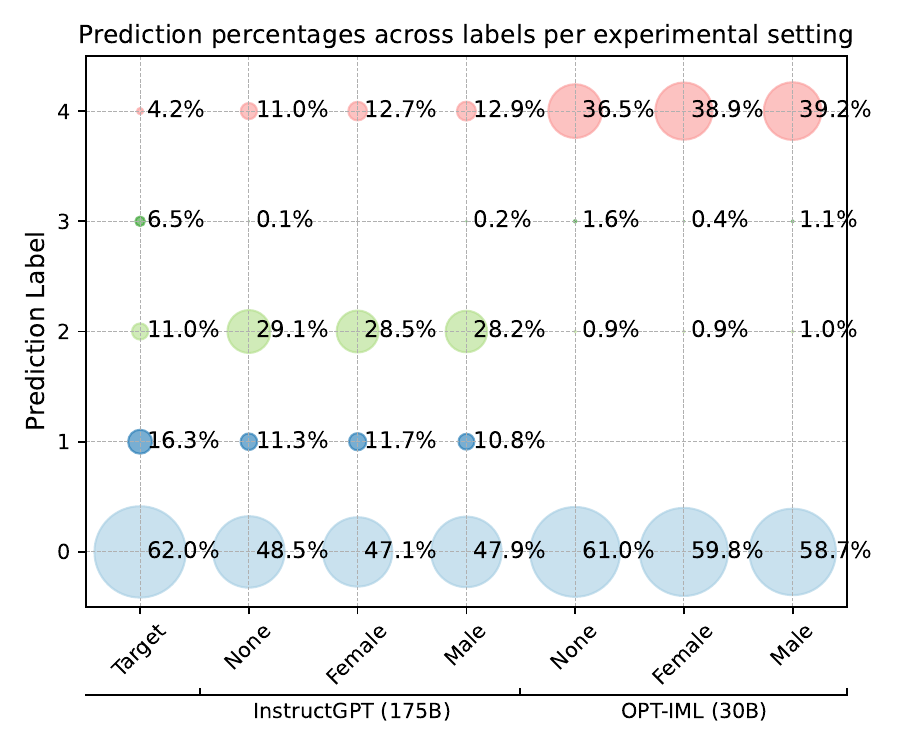}
    \caption{Prediction distribution across the different labels for the DP dataset. We compare the true label distribution (\texttt{Target}) with the results of different experimental settings for models \instructgpt (175B) and \imlopt (30B). \texttt{None} refers to prompting without sociodemographic information. \texttt{Female} and \texttt{Male} refer to sociodemographic prompting with a single attribute, respectively. The model choice has a larger influence on label predictions than the sociodemographic profile.}
    \label{fig:model_choice_influence}
\end{figure}

\paragraph{Increased model sensitivity does not translate to better performance.}

We test whether the percentage of prediction changes leads to better performance but do not measure any significant correlation (-0.16 Spearman $\rho$, p=0.08) when correlating performance improvement and percentage of prediction changes.
We conclude that model sensitivity is not a decisive factor for improvement of zero-shot performance using sociodemographic information. 

\paragraph{Model choice has large influence on label prediction.}

We established the influence of sociodemographic prompting on model performance but also observe a statistical significant effect of the model family.
On average, \texttt{InstructGPT} and \texttt{OPT-IML} variants perform best, while variants of \texttt{Flan}, \texttt{Tk-Instruct} and \texttt{Dolly-V2} perform significantly worse, independent of model size (§\ref{app:sec:zeroshot_performance}).
To better understand these differences, we visualize the percentage distribution of label predictions for different experimental settings in Figure~\ref{fig:model_choice_influence}.
Within the same model, we observe minor differences when changing the value of the sociodemographic attribute.
However, the majority of predictions are determined by the choice of model family.
Concretely, the predictions of \instructgpt are distributed differently across the label space than those of \imlopt. 
A similar picture emerges when the results are compared with other models (cf. §\ref{app:sec:model_choice_influence}).

To explain this observation, we investigate if our datasets (or parts thereof) are contained in the relevant instruction-tuning datasets and conclude that most models\footnote{The exact composition of the training datasets of \instructgpt and \dolly are disclosed.} have been exposed to the same tasks which are relevant for our datasets.
However, we note that \texttt{OPT-IML} was exposed to the largest number and variety of tasks and datasets during instruction fine-tuning (cf. Table~\ref{app:sec:modeldetails}).

\section{Robustness}
\label{ssec:robustness}

Previous work demonstrated that prompting for text classification is influenced by the prompt format~\citep{min-etal-2022-rethinking}.
Thus, we investigate whether sociodemographic prompting is \emph{robust} as indicated by the extent to which predictions change when reformulating the instruction.

\subsection{Detailed Setup} 
We compare the previous format used to two other formulations of the instruction; a paraphrase (1) and another where we do not provide any explicit instruction but merely present the sociodemographic profile and the input text (2).
We provide the exact formulations in  Appendix~\ref{app:sec:robustness}. 
Importantly, the sociodemographic profile remains the same between different formats.

\begin{table}
    \centering
    \resizebox{.49\textwidth}{!}{%
    \begin{tabular}{llclc}
    \toprule
    & \multicolumn{2}{c}{InstructGPT (175B)} & \multicolumn{2}{c}{OPT-IML (30B)} \\
         Model & Diff (1,2) & F1 & Diff (1,2) & F1 \\
    \midrule
    DP & (19\%, 33\%) & .27 \stdintable{.02} & (13\%, 82\%) & .15 \stdintable{.06} \\
    DP+SD & (10\%, 40\%) & .27 \stdintable{.02} & (15\%, 56\%) & \textbf{.20} \stdintable{.01} \\ 
     
    Jigsaw & (5\%, 16\%) & .61 \stdintable{.01} & (11\%, 30\%) & .50 \stdintable{.02} \\ 
    Jigsaw+SD & (4\%, 13\%) & .61 \stdintable{.01} & (14\%, 23\%) & \textbf{.53} \stdintable{.00} \\
     
    GHC & (2\%, 10\%) & .71 \stdintable{.04} & (6\%, 22\%) & .67 \stdintable{.05} \\ 
    GHC+SD & (2\%, 9\%) & \textbf{.73} \stdintable{.01} & (8\%, 16\%) & \textbf{.72} \stdintable{.04}  \\
    
    H-Twitter& (3\%, 12\%) & .54 \stdintable{.06} & (11\%, 91\%) & \textbf{.38} \stdintable{.22} \\
    H-Twitter+SD & (4\%, 8\%) & .54 \stdintable{.03} & (12\%, 95\%) &  .36 \stdintable{.25} \\
    
    SE2016 & (10\%, 41\%) & .38 \stdintable{.19} & (13\%, 28\%) & \textbf{.50} \stdintable{.08} \\
    SE2016+SD & (8\%, 17\%) & \textbf{.52} \stdintable{.01} & (12\%, 20\%) & .43  \stdintable{.12} \\
    
    GWSD & (7\%, 24\%) & .66 \stdintable{.04} & (13\%, 20\%) & \textbf{.41} \stdintable{.10} \\ 
    GWSD+SD & (12\%, 19\%) & \textbf{.67} \stdintable{.01} & (10\%, 10\%) & .34 \stdintable{.09} \\
    
    Diaz & (16\%, 45\%) & \textbf{.33} \stdintable{.03} & (24\%, 45\%) & .21 \stdintable{.04} \\
    Diaz+SD & (14\%, 36\%) & .31 \stdintable{.04} & (35\%, 42\%) & \textbf{.26} \stdintable{.02} \\
    
    \bottomrule
    \end{tabular}
    }
    \caption{Differences in terms of prediction changes and performance between different prompt formulations.
    Diff refers to prediction changes when comparing results of using format 0 to other formats (1,2).
    F1 refers to the averaged F1 scores across all three formats.
    }
    \label{tab:format_label_changes}
\end{table}

\subsection{Results}
\paragraph{Predictions are sensitive to prompt formulation.}

Table~\ref{tab:format_label_changes} presents the differences in prediction changes (in percent) between the different formats across datasets.
Even for semantically equivalent formulations (0,1) prediction differences can rise up to 35\% (\imlopt on \diaz{}).
Using a minimal format leads to the most drastic changes across all datasets, especially pronounced for  \divp{} and \htwitter{}.
Similar effects can be observed for prompting without sociodemographic information.
Thus, prediction differences are only partially induced by the sociodemographic profile and confirm previous observations that prompt formulation largely influences prediction outcomes.

\section{Discussion and Recommendations}
\label{ssec:discussion}

While all LLMs are sensitive to sociodemographic prompting, we identified model scale and the number of instruction-tuning tasks as relevant factors for improving model performance.
Toxicity detection and sentiment classification are the tasks which benefit the most from this technique.
Further, the model family and prompt formulation have a strong influence on model predictions.
Thus, we emphasize that sociodemographic prompting should be used with care, especially in human response simulation~\citep{durmus2023measuring} and data annotation~\citep{hartvigsen-etal-2022-toxigen}.

From these findings, we can extract actionable suggestions. 
First, estimating the degree of sociodemographic ``alignment'' of any LLM should not be merely based on the outcome of prompting with varying sociodemographic profiles. 
Our work points out the need of a general evaluation framework for studying the sociodemographic alignment of LLMs. 
Second, if any sociodemographic prompting experiment is conducted, a robustness analysis should accompany the work to evaluate the validity of the findings.

\paragraph{Sociodemographic prompting is effective at modeling disagreement.}

\begin{figure}[]
    \includegraphics[width=.49\textwidth]{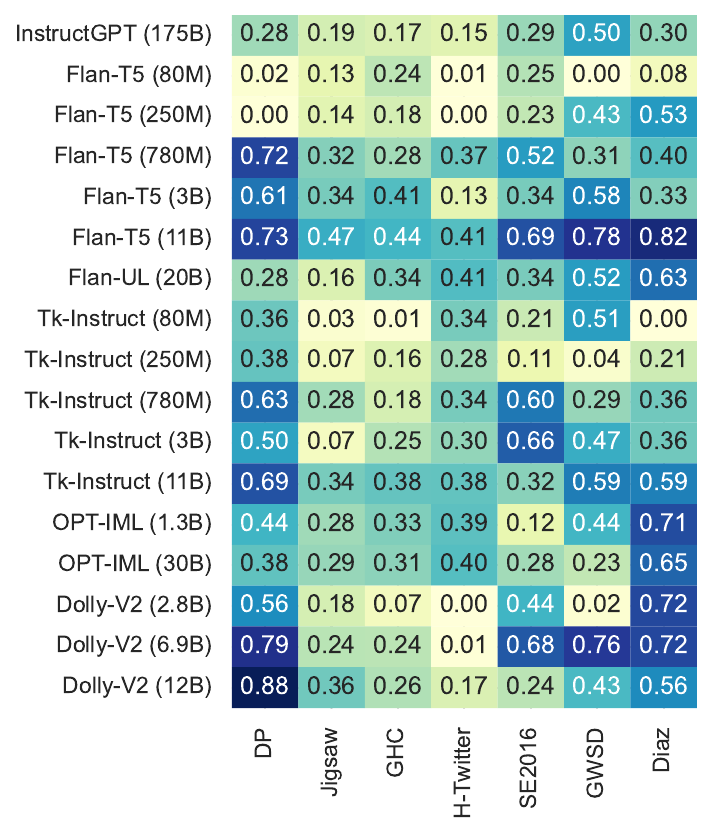}
    \caption{Performance to model disagreement in various datasets of subjective NLP tasks 
    (binary F1).
    }
    \label{fig:dispersion_heatmap}
\end{figure}

We also acknowledge potential applications of sociodemographic prompting in the future.
\citet{Wan_Kim_Kang_2023} trained a model for disagreement prediction in subjective NLP tasks.
Their approach relies on the existence of annotated data alongside the sociodemographic information of the annotators.
Thus, we investigate whether we can use sociodemographic prompting as an efficient method to identify instances which will likely result in disagreement during annotation.
We compare the original annotations with the result of sociodemographic prompting and calculate a binary F1 score. 
True positives are instances which received disagreement in both setups. 
Conversely, true negatives are instances which received no disagreeing votes in both setups.

We present the results in Figure~\ref{fig:dispersion_heatmap}.
Surprisingly, the best-performing zero-shot models (§\ref{ssec:zeroshot}) are not the best at modeling disagreement.
With a mean performance of 0.62, \flant (11B) produces the best and most consistent results across all datasets.
This is confirmed by our statistical analysis (§\ref{sec:glmm_analyses} for details).
For the two datasets (\divp{}, \diaz{}) with original sociodemographic information and lowest IAA overall, we observe the best performances across different model sizes.
As both datasets induce increased prediction changes (§\ref{ssec:model_sensitivity}), we hypothesize that sociodemographic prompting is more sensitive if there is larger disagreement in the original annotation. 
We interpret this as a promising result for using zero-shot sociodemographic prompting to estimate whether a text is likely to induce disagreement among annotators.
We leave the exploration of integrating additional training signals (e.g., few-shot) as future work.

\section{Conclusion}

We study sociodemographic prompting for subjective NLP tasks and employ a comprehensive study across seven datasets and seven instruction-tuned LLMs from different model families.
Our results show that these models are sensitive to sociodemographic prompting and using this technique can improve zero-shot performance.

However, we also observe a strong influence of the prompt formulation and model family.
Thus, we argue that sociodemographic prompting should be used with care in sensitive applications and requires comprehensive evaluation when used for data annotation or studying sociodemographic alignment of language models.

\section*{Limitations}

In the following, we provide an examination of the inherent limitations associated with this research study.
We further note that all our experiments have been approved by the local ethics review process of the Business School of the University of Hamburg. 
This process is compliant to obtaining approval from an Institutional Review Board.

\paragraph{Annotations go beyond sociodemographics.}

While annotators' sociodemographic backgrounds have been shown to be influential in their decision-making process~\citep[\emph{inter alia}]{al-kuwatly-etal-2020-identifying, excell-al-moubayed-2021-towards, shen-rose-2021-sounds, larimore-etal-2021-reconsidering, sap-etal-2022-annotators}, it is not a definitive predictive factor as individual lived experiences~\citep{waseem-2016-racist} or situated domain expertise~\citep{patton2019annotating} can influence annotation decisions, too.
In short, collective group behavior may not always provide an explanation for individual behavior~\citep{crowdworksheets_diaz2022, orlikowski-etal-2023-ecological}.
While our general approach can be extended to a wider range of sociodemographic attributes or even descriptions of individuals, we refrained from testing more to contain the complexity of our study and due to the limited availability of such resources.
We welcome efforts to increase the availability of such information alongside the datasets, e.g., Crowdworksheets by \citet{crowdworksheets_diaz2022}, and hope to see more work in future exploring prompting large language models with more dimensions of sociodemographic and personal information. 

\paragraph{Sociodemographic profiles are not representative.}

It is important to acknowledge certain limitations with regard to the representation of sociodemographic profiles. 
First, all the datasets employed in our research are exclusively in English language, mostly due to the lack of resources in other languages.
This linguistic restriction inherently limits our ability to make comprehensive cross-linguistic assessments. 
Second, the sociodemographic information provided by the annotators of the datasets used in this study adheres to a classification system specific to the United States. 
Consequently, our findings cannot be generalized to sociodemographic data originating from other nations, linguistic communities, or cultural contexts. 
These limitations underscore the need for caution when extrapolating our results to broader sociodemographic contexts beyond the scope of our study.

\paragraph{We cannot model all factors influencing prompting outcomes.}

We demonstrate that model predictions can effectively be changed when incorporating sociodemographic information within the prompt (§\ref{ssec:model_sensitivity}).
However, we acknowledge that this is one among many of the factors influencing model predictions in a zero-shot prompting setup.
We account for the influence of prompt formulation by investigating its effect in §\ref{ssec:robustness} and are aware of the growing body of work investigating various other factors which influence prompting results, such as correct label assignment~\citep{min-etal-2022-rethinking}, domain-specific vocabulary~\citep{fei-etal-2023-mitigating} or example order~\citep{kumar-talukdar-2021-reordering, pmlr-v139-zhao21c, lu-etal-2022-fantastically}.

The majority of these works deals with in-context learning or few-shot learning in general which we do not investigate in this study.
However, we see these phenomena as support for our overall argument (§\ref{ssec:robustness}) that estimating the degree of alignment of any LLM should not be merely based on the outcome of prompting with varying sociodemographic profiles.
This is due to their lack of robustness when changing the surface form of the prompt while keeping its semantic meaning similar.

\paragraph{Simulating annotations using prompting mechanisms is limited.}

Employing humans for annotation projects in NLP is a multi-step process.
It involves the formulation of annotation guidelines and their iterative refinement through discussions between the annotators and coordinators.
In most cases, annotators are undergoing a qualification process or test to evaluate their eligibility for contributing to the annotations. 
These factors influence the decision-making process of the annotators and ultimately the annotation agreement.
In our experimental setup, we do not provide any additional instructions to the model than the prompt instructions which we present in §\ref{sec:sociodemographic_prompting} and §\ref{app:sec:robustness}, thus possibly underspecifying the task instruction to the model.
Our experimental setup is designed driven by the following observations;
for most datasets, the original annotation guidelines are non-retrievable and could only be guessed from the description in the corresponding research publication.
Further, LLMs are limited with regards to the context input size (see Table~\ref{app:sec:modeldetails} for details) and using longer prompts would have limited our experiments to a few models with appropriate input sizes. 

\section*{Acknowledgements}

We thank Max Glockner, Hovhannes Tamoyan and Anmoel Goel for their feedback on an early draft of this work and the authors of the datasets we used for providing them publicly.
We gratefully acknowledge the support of Microsoft with a grant for access to OpenAI GPT models via the Azure cloud (Accelerate Foundation Model Academic Research).
This work has been funded by the German Federal Ministry of Education and Research (BMBF) under the promotional reference 01UP2229B (KoPoCoV) and by the German Research Foundation (DFG) as part of the Research Training Group KRITIS No. GRK 2222. 
Anne Lauscher’s work is funded under the Excellence Strategy of the Federal Government and the Länder.

\bibliography{anthology,custom}
\bibliographystyle{acl_natbib}

\clearpage

\appendix
\section{Appendix}

\subsection{Dataset details}
\label{app:sec:datasets}

\paragraph{Toxicity - DP.}
The \divp~dataset comprises comments extracted from various online forums, including Twitter, 4chan, and Reddit, spanning from December 2019 to August 2020.
These comments underwent annotation via Amazon Mechanical Turk, receiving five annotations per instance. 
Sociodemographic data of the annotators was gathered, contingent upon the approval of the pertinent Institutional Review Board (IRB). 
The dataset did not come equipped with a definitive gold label. 
Therefore, we instituted a majority voting mechanism, leveraging the raw annotations to ascertain the gold label.
We exclude instances wherein selected sociodemographic attributes received responses such as "Prefer not to say", instances with multiple selections for the \textit{race} attribute, and any  attributes marked with generic designations like \textit{Other} or \textit{Unknown}.
This reduced the initial dataset size from 107,620 to 55,364 instances.

\paragraph{Toxicity - Jigsaw.}
We filtered all instances where unaggregated annotations were missing, which reduced the dataset size from 1,999,516 to 1,804,874 instances.

\paragraph{Hatespech - GHC.}
We filtered all instances where no gold annotation was provided.
Thus, the initial dataset size was reduced from 27,553 to 27,434.

\paragraph{Hatespeech - H-Twitter.}

For H-Twitter, none of 6,909 instances were filtered.

\paragraph{Stance Detection - SE2016.}
We only considered instances were both the aggregated and the complete set of unaggregated annotations were available.
In addition, we removed the hashtag '\#SemST' which was artificially added by the dataset authors.
The dataset we used consists of 3,591 instances.

\paragraph{Stance - GWSD.}
It is composed of a subset of 2,050 annotated articles, extracted from a larger pool of 56,000 articles on global warming. 
These articles were published between January 1, 2000, and April 12, 2020, by 63 distinct U.S. news outlets. 
After filtering instances where no annotations were provided and removing duplicates, 2,042 instances remained of the initial 2,050.

\paragraph{Sentiment - Diaz.}

The Diaz dataset is the second dataset where sociodemographic data of the annotators was gathered.
We only considered the sociodemographic attributes which were also used in the \divp~dataset.
To remain comparability, we convert the original 5-point answer for political affiliations into a 3-scale by mapping 'Somewhat conservative' and 'Very conservative' to 'Conservative' and 'Somewhat liberal' and 'Very liberal' to 'Liberal'.
We filtered all instances where annotators replied with \textit{Other} for the attribute race.
The dataset which was used for sampling contains 14,071 instances.

\subsection{Sampling Strategy}
\label{app:sec:sampling}

\begin{table*}
    \centering
    \resizebox{1.00\textwidth}{!}{%
    \begin{tabular}{l|l|l|c}
        \toprule
        \textbf{Sociodemographic Attribute} & \textbf{Value} & \textbf{Original} & \textbf{Sample (n=1,000)}  \\
       \midrule
       Gender & female & 52.18 & 52.56 \\
        & male & 47.35 & 46.88  \\
        & nonbinary & 00.47 & 00.56  \\
        \midrule
       Race & White & 76.90 & 77.62 \\
       & Black/African American & 13.12 & 12.40) \\
       & Asian & 6.15 & 6.34 \\
       & Hispanic & 2.78 & 2.62 \\
       & American Indian or Alaska Native & 00.80 & 0.90 \\
       & Native Hawaiian or Pacific Islander & 00.24 & 00.12 \\ 
        \midrule
       Education & Bachelor's degree in college (4-year) & 41.87 & 42.58 \\
       & Some college but no degree & 19.18 & 18.38 \\
       & Master's degree & 15.90 & 15.90 \\ 
       & Associate degree in college (2-year) & 10.93 & 10.90 \\ 
       & High school graduate (high school diploma or equivalent including GED) & 8.69 & 8.76 \\ 
       & Professional degree (JD, MD) & 1.59 & 1.80 \\
       & Doctoral degree & 1.32 & 1.14 \\
       & Less than high school degree & 00.53 &  00.54 \\ 
       \midrule
       Age Range & 25 - 34 & 39.62 & 39.00 \\
       & 35 - 44  & 25.03 & 25.08 \\ 
       & 45 - 54 & 13.49 & 13.66 \\ 
       & 18 - 24 & 10.73 & 10.92 \\ 
       & 55 - 64 & 7.70 & 8.08 \\ 
       & 65 or older & 3.41 & 3.26 \\
       & Under 18 & 00.02 & 00.00 \\
       \midrule
       Political Affiliation & Liberal & 43.04 & 43.70 \\ 
       & Conservative & 28.77 & 28.80 \\ 
       & Independent & 28.18 & 27.50 \\ 
       \bottomrule
    \end{tabular}}
    \caption{Distribution of sociodemographic attributes for both the full dataset and the sample of \divp{}.}
    \label{app:tab:sampling_distribution}
\end{table*}

Due to the large number of experiments and varying dataset sizes, we first randomly sample 1,000 instances from each dataset.
Here, the label distribution of the samples remained comparable to the corresponding full dataset distribution. 
To conduct our sociodemographic prompting experiments, we used the original sociodemographic profiles for \divp{} and \emph{Diaz} datasets while transferring the sociodemographic profiles of the \divp{} samples to the remaining datasets (where no sociodemographic info about the annotators is provided). 
Thus, the sociodemographic profiles (and their respective distribution) which we use for datasets \divp{}, \jigsaw{}, \ghc{}, \htwitter{}, \semeval{}, and \gwsd{} are identical (five per instance). 
We display their distribution (Table~\ref{app:tab:sampling_distribution}) and compare it to the distribution of the full \divp{} dataset.
As can be seen from the table, both remain comparable.

\subsection{Model Details}
\label{app:sec:modeldetails}

\begin{table}
    \centering
    \resizebox{.49\textwidth}{!}{%
    \begin{tabular}{cccc}
        \toprule
        \textbf{Model} & \textbf{Parameters} & \textbf{Context Size} & \textbf{Nr Tasks} \\
       \midrule
       \instructgpt & 175B & 4097 & - \\
       \midrule
        \flant & \href{https://huggingface.co/google/flan-t5-small}{80M}  & 512 & 1,836 \\
               & \href{https://huggingface.co/google/flan-t5-base}{250M} & 512 & \\
               & \href{https://huggingface.co/google/flan-t5-large}{780M} & 512 & \\
               & \href{https://huggingface.co/google/flan-t5-xl}{3B} & 512 & \\
               & \href{https://huggingface.co/google/flan-t5-xxl}{11B} & 512 & \\
        \midrule
        \flantul & \href{https://huggingface.co/google/flan-ul2}{20B} & 512 & 1,836 \\
        \midrule
        \tkinstruct & \href{https://huggingface.co/allenai/tk-instruct-small-def-pos}{80M} & 512 & 1,616 \\
               & \href{https://huggingface.co/allenai/tk-instruct-base-def-pos}{250M} & 512 & \\
               & \href{https://huggingface.co/allenai/tk-instruct-large-def-pos}{780M} & 512 & \\
               & \href{https://huggingface.co/allenai/tk-instruct-3b-def-pos}{3B} & 512 & \\
               & \href{https://huggingface.co/allenai/tk-instruct-11b-def-pos}{11B} & 512 & \\
        \midrule
        \imlopt & \href{https://huggingface.co/facebook/opt-iml-1.3b}{1.3B} & 2048 & 1,991 \\
                & \href{https://huggingface.co/facebook/opt-iml-30b}{30B} & 2048 & \\
        \midrule
        \dolly & \href{https://huggingface.co/databricks/dolly-v2-3b}{2.8B} & 2560 & 7\\
                & \href{https://huggingface.co/databricks/dolly-v2-7b}{6.9B} & 4096 & \\
                & \href{https://huggingface.co/databricks/dolly-v2-12b}{12B} & 5120 & \\
        \bottomrule 
    \end{tabular}}
    \caption{The models and configurations we use. The last column indicates the number of instruction-tuning tasks which were used to train the model.}
    \label{app:tab:models}
\end{table}

We provide an overview of all models, their size in terms of number of parameters and their context window size in Table~\ref{app:tab:models}. 
For models with parameters less than 3B (i.e. Flan-T5 80M-3B, Tk-Instruct 80M-3B, OPT-IML 3B and Dolly-V2 2.8B) no inference optimization was applied.
We used 8-bit optimization for all other models.

\paragraph{Licenses} Models from \flant, \flantul, and \tkinstruct are published using an Apache License 2.0. \imlopt-based models are published with the OPT-IML 175B LICENSE AGREEMENT which allows usage for non-commercial research purposes.
\dolly is distributed under MIT license.

\subsection{Experimental Details}
\label{app:sec:experimental_details}

All experiments were conducted using PyTorch~\citep{Paszke2019PyTorchAI} 2.0.1, Huggingface Transformers~\citep{Wolf2019HuggingFacesTS} 4.28.1 and CUDA~\citep{nickolls2008scalable} 11.8 on a computation cluster with a combination of A100 (40GB), A180 (80GB) and H100PCIE (80GB) GPU cards.
Depending on the dataset and the GPU in use, we used batch sizes ranging from 4 to 32.
We used 8-bit optimization~\citep{dettmers2022optimizers} for models with parameter numbers larger than 3B.
For prompting \instructgpt, we used both the \href{https://platform.openai.com/docs/api-reference}{OpenAI API} and \href{https://azure.microsoft.com/en-us/products/ai-services/openai-service}{Microsoft Azure API}.

\subsection{Model Sensitivity}
\label{app:sec:model_sensitivity}

Here, we provide more details regarding the experiments to analyze the sensitivity of instruction-tuned language models when prompted with (or without) sociodemographic information.

\subsubsection{Prediction changes per model and dataset}
\label{app:sec:label_changes_per_model_and_dataset}

\begin{figure*}
    \centering
    \includegraphics[width=1.0\textwidth]{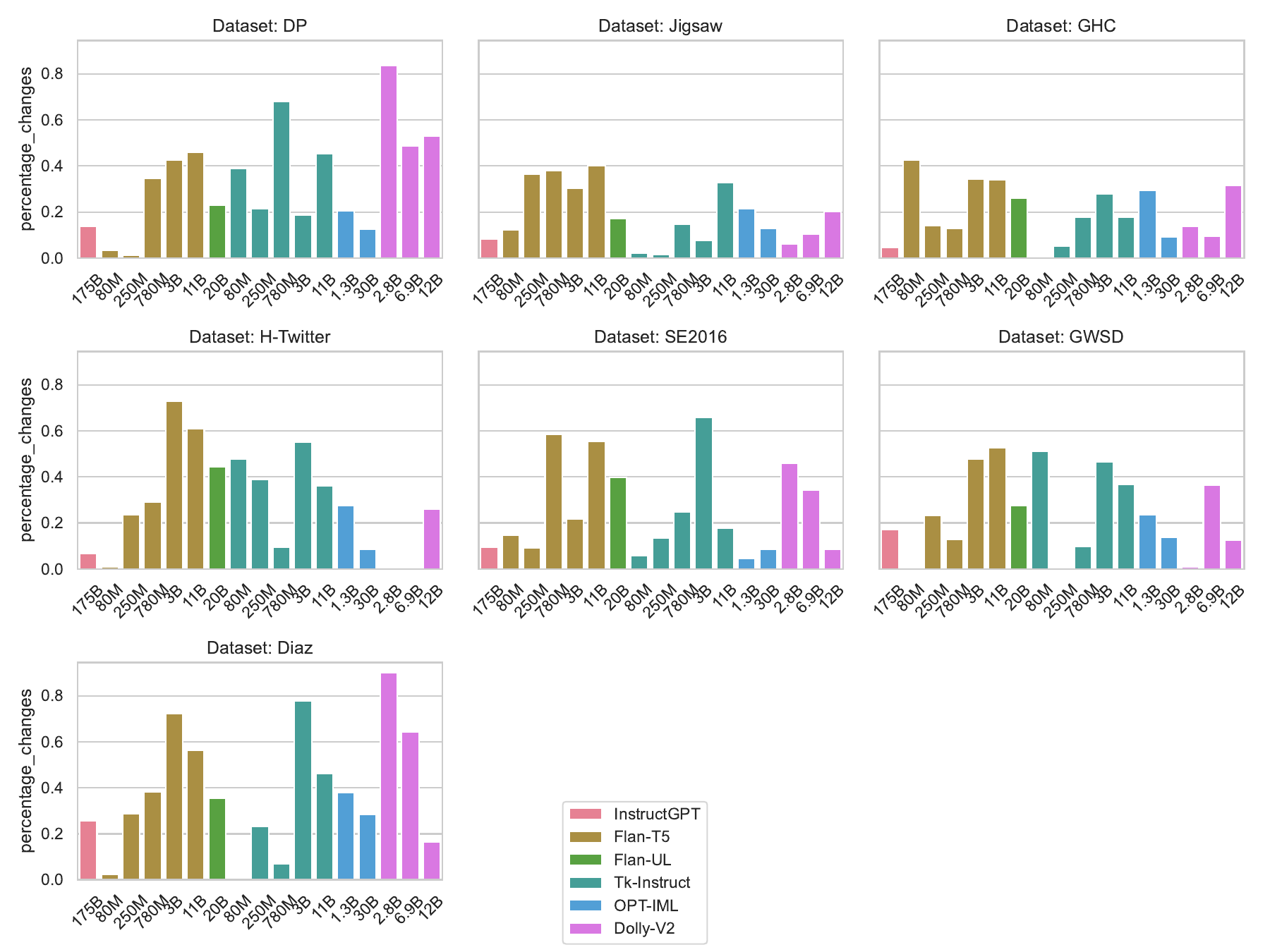}
    \caption{Percentage of prediction changes when comparing outputs of zero-shot prompting with and without sociodemographic information. The x-axis displays the model sizes of various instruction-tuned model families (same color).}
    \label{app:fig:label_changes}
\end{figure*}

In Figure~\ref{app:fig:label_changes} we display the degree of prediction changes for the different models and datasets.

\subsubsection{Prediction changes per sociodemographic attribute}
\label{app:sec:label_changes_per_attribute}

For better overview of the attributes which lead to the most prediction changes, we aggregated the most influential attributes in Table~\ref{app:tab:label_changes_per_attribute_aggregated}.
It can be seen that a combination of all sociodemographic attributes leads to the most substantial changes for most of the datasets (64\%).

\begin{table*}[]
    \centering
     \resizebox{1.00\textwidth}{!}{%
    \begin{tabular}{llllllll}
    \toprule
    Model (Params) & DP & Jigsaw & GHC & H-Twitter & SE2016 & GWSD & Diaz \\
    \midrule
\instructgpt (175B) & \cellcolor{R}R (17.30\%) & \cellcolor{All}All (8.20\%) & \cellcolor{R}R (5.50\%) & \cellcolor{R}R (7.90\%) & \cellcolor{All}All (9.60\%) & \cellcolor{All}All (17.12\%) &  \cellcolor{PA}PA (36.60\%) \\
\flant (80M) & \cellcolor{All}All (3.20\%) & \cellcolor{All}All (12.30\%) & \cellcolor{All}All (42.50\%) &  \cellcolor{PA}PA (2.60\%) & \cellcolor{All}All (14.80\%) & \cellcolor{All}All (0.00\%) & \cellcolor{All}All (2.30\%) \\
\flant (250M) & \cellcolor{A}A (1.80\%) & \cellcolor{All}All (36.40\%) & \cellcolor{R}R (16.40\%) & \cellcolor{All}All (23.60\%) & \cellcolor{All}All (9.10\%) &  \cellcolor{E}E (33.93\%) & \cellcolor{All}All (28.60\%) \\
\flant (780M) & \cellcolor{All}All (34.70\%) & \cellcolor{All}All (38.00\%) & \cellcolor{R}R (17.30\%) & \cellcolor{All}All (29.10\%) & \cellcolor{All}All (58.30\%) & \cellcolor{All}All (12.91\%) & \cellcolor{All}All (38.20\%) \\
\flant (3B) &  \cellcolor{PA}PA (46.80\%) & \cellcolor{R}R (51.90\%) &  \cellcolor{E}E (38.10\%) & \cellcolor{All}All (72.70\%) & \cellcolor{All}All (21.70\%) & \cellcolor{All}All (47.85\%) & \cellcolor{All}All (72.30\%) \\
\flant (11B) &  \cellcolor{PA}PA (50.40\%) & \cellcolor{R}R (41.30\%) & \cellcolor{R}R (36.70\%) & \cellcolor{R}R (64.90\%) & \cellcolor{All}All (55.40\%) & \cellcolor{All}All (52.75\%) & \cellcolor{R}R (64.00\%) \\
\flantul (20B) &  \cellcolor{E}E (25.60\%) &  \cellcolor{G}G (25.20\%) & \cellcolor{R}R (32.80\%) & \cellcolor{R}R (56.60\%) & \cellcolor{All}All (39.70\%) & \cellcolor{All}All (27.63\%) & \cellcolor{All}All (35.50\%) \\
\tkinstruct (80M) & \cellcolor{All}All (39.00\%) & \cellcolor{All}All (2.00\%) & \cellcolor{All}All (0.60\%) & \cellcolor{R}R (58.40\%) & \cellcolor{All}All (5.70\%) & \cellcolor{All}All (50.95\%) & \cellcolor{All}All (0.00\%) \\
\tkinstruct (250M) & \cellcolor{All}All (21.40\%) & \cellcolor{All}All (1.60\%) & \cellcolor{All}All (5.20\%) & \cellcolor{All}All (39.00\%) & \cellcolor{All}All (13.50\%) & \cellcolor{R}R (0.80\%) & \cellcolor{All}All (23.30\%) \\
\tkinstruct (780M) & \cellcolor{All}All (67.90\%) &  \cellcolor{PA}PA (14.80\%) & \cellcolor{All}All (17.70\%) & \cellcolor{All}All (9.40\%) & \cellcolor{All}All (24.70\%) &  \cellcolor{E}E (12.61\%) & \cellcolor{All}All (7.00\%) \\
\tkinstruct (3B) & \cellcolor{All}All (18.80\%) &  \cellcolor{PA}PA (7.90\%) & \cellcolor{All}All (27.80\%) & \cellcolor{All}All (55.00\%) & \cellcolor{All}All (65.70\%) & \cellcolor{All}All (46.55\%) & \cellcolor{All}All (77.70\%) \\
\tkinstruct (11B) & \cellcolor{All}All (45.20\%) & \cellcolor{All}All (32.80\%) & \cellcolor{R}R (19.20\%) & \cellcolor{All}All (36.00\%) & \cellcolor{All}All (17.70\%) & \cellcolor{All}All (36.74\%) & \cellcolor{All}All (46.20\%) \\
\imlopt (1.3B) &  \cellcolor{PA}PA (22.00\%) & \cellcolor{R}R (30.00\%) & \cellcolor{R}R (34.20\%) & \cellcolor{All}All (27.60\%) &  \cellcolor{G}G (4.70\%) & \cellcolor{All}All (23.42\%) & \cellcolor{All}All (37.90\%) \\
\imlopt (30B) & \cellcolor{All}All (12.50\%) & \cellcolor{R}R (12.90\%) & \cellcolor{All}All (9.30\%) & \cellcolor{R}R (9.90\%) &  \cellcolor{E}E (8.80\%) & \cellcolor{R}R (14.61\%) & \cellcolor{All}All (28.50\%) \\
\dolly (2.8B) & \cellcolor{All}All (83.60\%) & \cellcolor{All}All (6.20\%) & \cellcolor{All}All (13.60\%) & \cellcolor{All}All (0.40\%) & \cellcolor{All}All (46.00\%) & \cellcolor{All}All (0.90\%) & \cellcolor{All}All (89.90\%) \\
\dolly (6.9B) & \cellcolor{All}All (48.60\%) & \cellcolor{All}All (10.40\%) & \cellcolor{All}All (9.50\%) & \cellcolor{All}All (0.30\%) &  \cellcolor{G}G (39.10\%) &  \cellcolor{PA}PA (53.65\%) &  \cellcolor{PA}PA (64.90\%) \\
\dolly (12B) &  \cellcolor{E}E (59.80\%) & \cellcolor{All}All (20.30\%) & \cellcolor{A}A (35.80\%) &  \cellcolor{PA}PA (29.80\%) &  \cellcolor{G}G (16.20\%) & \cellcolor{A}A (14.61\%) &  \cellcolor{PA}PA (19.60\%) \\
    \bottomrule
    \end{tabular}
    }
    \caption{Most influential sociodemographic attribute per model and dataset, and in brackets the percentage of label changes due to sociodemographic prompting when compared to prompting without any sociodemographic information. \colorbox{All}{All} refers to the combination of all sociodemographic attributes, \colorbox{PA}{PA} refers to \textit{Political Affiliation}, \colorbox{R}{R} refers to \textit{Race}, \colorbox{A}{A} refers to \textit{Age-Range}, \colorbox{E}{E} refers to \textit{Education} and \colorbox{G}{G} refers to \textit{Gender}.}
    \label{app:tab:label_changes_per_attribute_aggregated}
\end{table*}

\subsection{Zero-Shot Prompting using sociodemographic information}
\label{app:sec:zeroshot_performance}

\subsubsection{Predicting annotator-specific annotations}

As extension to the results provided in §\ref{ssec:zeroshot}, in Table~\ref{app:tab:zeroshot_original_annotations} we provide the results for all models when predicting an annotator-specific annotation with the same sociodemographic profile as provided in the prompt. 

The largest models in our experiments (20B, 30B, 175B) all benefit from integrating the sociodemographic information provided with the original annotation.
However, for smaller models there is no consistent trend of improvement observable.
Interestingly, most models from instruction-tuned model families based on T5 (\flant, \tkinstruct) and Pythia (\dolly) are not able to outperform random guessing.
This is independent of the model size.

\begin{table}
    \centering
    \resizebox{.49\textwidth}{!}{%
    \begin{tabular}{l|cc|cc}
    \toprule
    & \multicolumn{2}{c}{Toxicity - DP} & \multicolumn{2}{c}{Sentiment - Diaz} \\
         Model & Acc & F1 & Acc & F1 \\
    \midrule
    Random & .19 & .17 & .20 & .17 \\
    Majority & .06 & .02 & .09 & .03 \\
    \midrule
    \instructgpt(175B) & .43 & .26 & .34 & .26 \\
    \instructgpt(175B)+SD & \textbf{.44} & .26 & \textbf{.37} & \textbf{.31} \\
    \midrule
    \flant(80M) & .50 & \textbf{.15} & .09 & .04 \\
    \flant(80M)+SD & \textbf{.51} & .14 & .09 & .04 \\
    \flant(250M) & \textbf{.20} & .07 & .15 & .10 \\
    \flant(250M)+SD & .19 & .07 & \textbf{.17} & \textbf{.11} \\
    \flant(780M) & \textbf{.20} & .09 & .20 & .10 \\
    \flant(780M)+SD & .18 & \textbf{.11} & \textbf{.33} & \textbf{.13} \\
    \flant(3B) & \textbf{.51} & .14 & .12 & .07 \\
    \flant(3B)+SD & .30 & \textbf{.15} & \textbf{.36} & \textbf{.13} \\
    \flant(11B) & \textbf{.25} & \textbf{.15} & \textbf{.29} & .17 \\
    \flant(11B)+SD & .19 & .12 & .26 & .17 \\
    \midrule
    \flantul(20B) & .36 & \textbf{.16} & .08 & .06 \\
    \flantul(20B)+SD & \textbf{.40} & .14 & \textbf{.15} & \textbf{.11} \\
    \midrule
    \tkinstruct(80M) & \textbf{.18} & .08 & .05 & .02 \\
    \tkinstruct(80M)+SD & .12 & .08 & .05 & .02 \\
    \tkinstruct(250M) & .18 & .11 & \textbf{.05} & .02 \\
    \tkinstruct(250M)+SD & \textbf{.22} & .11 & .04 & \textbf{.03} \\
    \tkinstruct(780M) & \textbf{.42} & \textbf{.18} & .05 & .02 \\
    \tkinstruct(780M)+SD & .19 & .13 & .05 & \textbf{.03} \\
    \tkinstruct(3B) & \textbf{.49} & .16 & .15 & .12 \\
    \tkinstruct(3B)+SD & .40 & .16 & \textbf{.38} & \textbf{.15} \\
    \tkinstruct(11B) & \textbf{.18} & .11 & .11 & .10 \\
    \tkinstruct(11B)+SD & .15 & \textbf{.12} & \textbf{.14} & \textbf{.13} \\
    \midrule
    \imlopt(1.3B) & .44 & .18  & \textbf{.28} & \textbf{.24} \\
    \imlopt(1.3B)+SD & \textbf{.48} & .18 & .27  & .22 \\
    \imlopt(30B) & .42 & .18 & .28 & .26 \\
    \imlopt(30B)+SD & \textbf{.45} & .18 & \textbf{.32} & \textbf{.27} \\
    \midrule
    \dolly (2.8B) & \textbf{.29} & \textbf{.16} & .07 & .07  \\
    \dolly (2.8B)+SD & .12 & .09  & \textbf{.15}  & \textbf{.13} \\
    \dolly (6.9B) & \textbf{.43} & .16 & \textbf{.21}  & \textbf{.21}  \\
    \dolly (6.9B)+SD & .27 & .16  & .11  & .16  \\
    \dolly (12B) & .09 & .08 & .26 & .13 \\
    \dolly (12B)+SD & \textbf{.12} & \textbf{.12}  & .26  & \textbf{.15}  \\
    \bottomrule
    \end{tabular}
    }
    \caption{Zero-shot performance when predicting annotator-specific annotations using the original sociodemographic profile. We compare prompting with (SD) and without sociodemographic information and report macro-averaged F1 and Accuracy (Acc). Bold scores highlight the better performance when comparing the same model.}
    \label{app:tab:zeroshot_original_annotations}
\end{table}

\subsubsection{Predicting aggregated annotations}

The complete list of results for zero-shot prompting using sociodemographic profiles is provided in Table~\ref{app:tab:zeroshot_hard_all} (hard evaluation) and Table~\ref{app:tab:zeroshot_soft_all} (soft evaluation).
The good performance of models from the \tkinstruct family on the Jigsaw dataset is most likely due to the dataset being present in the dataset which was used for instruction-finetuning ~\citep{wang-etal-2022-super}.
The authors report toxic language detection with 40 datasets being one of the most prominent tasks among all.
Some of the results for GWSD can be explained in a similar vein as stance detection has been part of the instruction-tuning tasks present in the dataset.

\begin{table*}[]
    \centering
    \resizebox{1.00\textwidth}{!}{%
    \begin{tabular}{lc|cc|cc|cc|cc|cc|cc|cc|cc|cc|cc|cc|cc|cc|cc}
    \toprule
         & & \multicolumn{4}{c}{Toxicity} & \multicolumn{4}{c}{Hatespeech} & \multicolumn{4}{c}{Stance Detection} & \multicolumn{2}{c}{Sentiment}  \\
         & & \multicolumn{2}{c}{DP} & \multicolumn{2}{c}{Jigsaw} & \multicolumn{2}{c}{GHC} & \multicolumn{2}{c}{H-Twitter} & \multicolumn{2}{c}{SE2016} & \multicolumn{2}{c}{GWSD} & \multicolumn{2}{c}{Diaz} \\
         Model & Avg & Acc & F1 & Acc & F1 & Acc & F1 & Acc &F1 & Acc &F1 & Acc &F1 & Acc & F1 \\
         \midrule
InstructGPT(175B)&.66 &.48&.27&.76&.59&.89&\textbf{.75}&\textbf{.87}&\textbf{.57}&\textbf{.53}&.51&\textbf{.72}&\textbf{.69}&.36&.33\\
InstructGPT(175B)+SD&.66&\textbf{.51}&\textbf{.28}&\textbf{.79}&\textbf{.60}&.89&.74&.86&.53&.52&\textbf{.52}&.69&.68&\textbf{.39}&.33\\
\midrule
Flan-T5(80M)&.27&.61&\textbf{.17}&.21&.21&.32&.31&.01&.00&.21&.13&.45&.21&.06&.03\\
Flan-T5(80M)+SD&\textbf{.33}&.61&.16&\textbf{.32}&\textbf{.30}&\textbf{.64}&\textbf{.48}&.01&\textbf{.01}&\textbf{.24}&\textbf{.17}&.45&.21&.06&.03\\
Flan-T5(250M)&\textbf{.34}&\textbf{.17}&.06&\textbf{.40}&\textbf{.33}&.19&.19&.63&.20&\textbf{.56}&.27&\textbf{.27}&.21&.13&.08\\
Flan-T5(250M)+SD&.33&.16&.06&.12&.12&\textbf{.22}&\textbf{.22}&\textbf{.84}&\textbf{.23}&.54&\textbf{.28}&.25&.21&\textbf{.15}&\textbf{.09}\\
Flan-T5(780M)& \textbf{.29}&\textbf{.18}&.09&\textbf{.54}&\textbf{.38}&\textbf{.29}&\textbf{.26}&.05&.04&\textbf{.37}&\textbf{.26}&\textbf{.42}&.21&.20&.10\\
Flan-T5(780M)+SD&.24&.16&.09&.22&.21&.24&.22&\textbf{.09}&\textbf{.07}&.24&.21&.40&\textbf{.23}&\textbf{.36}&\textbf{.13}\\
Flan-T5(3B)&\textbf{.34}&\textbf{.61}&\textbf{.16}&\textbf{.67}&\textbf{.54}&\textbf{.37}&\textbf{.35}&\textbf{.09}&\textbf{.06}&\textbf{.26}&\textbf{.22}&.30&.25&.10&.06\\
Flan-T5(3B)+SD&.32&.35&.15&.55&.45&.21&.19&.02&.01&.23&.20&\textbf{.46}&\textbf{.37}&\textbf{.42}&\textbf{.13}\\
Flan-T5(11B)&\textbf{.38}&\textbf{.23}&\textbf{.14}&\textbf{.42}&\textbf{.34}&\textbf{.38}&\textbf{.33}&\textbf{.67}&\textbf{.21}&.30&\textbf{.26}&.36&.27&\textbf{.32}&.16\\
Flan-T5(11B)+SD&.25&.17&.10&.13&.13&.16&.16&.27&.12&\textbf{.33}&.24&\textbf{.40}&\textbf{.33}&.28&.16\\
\midrule
Flan-UL(20B)&.29& .43 & \textbf{.17} & .63 & .39 & \textbf{.39} & \textbf{.30} & .01 & .01 & .15 & \textbf{.15} & .34 & \textbf{.20} & .07 & .05 \\
Flan-UL(20B)+SD&.29& \textbf{.50} & .14 & \textbf{.64} & .39 & .21 & .18 & \textbf{.03} & \textbf{.03} & \textbf{.16} & .11 & .34 & .19 & \textbf{.16} & \textbf{.10} \\
\midrule
Tk-Instruct(80M)&\textbf{.52}&\textbf{.15}&\textbf{.07}&\textbf{.90}&.48&\textbf{.87}&.46&\textbf{.70}&\textbf{.24}&\textbf{.58}&.25&\textbf{.44}&.23&.03&.01\\
Tk-Instruct(80M)+SD&.46&.10&.06&.88&\textbf{.49}&.86&.46&.41&.17&.56&\textbf{.26}&.41&\textbf{.30}&.03&.01\\
Tk-Instruct(250M)&\textbf{.46}&.17&.11&.90&.49&\textbf{.85}&.47&\textbf{.66}&\textbf{.21}&\textbf{.25}&\textbf{.18}&.35&.17&.03&.02\\
Tk-Instruct(250M)+SD&.43&\textbf{.23}&.11&\textbf{.91}&\textbf{.51}&.82&\textbf{.49}&.45&.16&.21&.13&\textbf{.36}&\textbf{.18}&.03&.02\\
Tk-Instruct(780M)&\textbf{.48}&\textbf{.48}&\textbf{.17}&.71&.50&.72&\textbf{.50}&.77&.23&\textbf{.22}&\textbf{.18}&.43&\textbf{.26}&.03&.01\\
Tk-Instruct(780M)+SD&.46&.17&.11&\textbf{.77}&.50&\textbf{.81}&.48&\textbf{.80}&.23&.20&.17&.43&.24&.03&\textbf{.02}\\
Tk-Instruct(3B)&.44&\textbf{.59}&\textbf{.17}&.83&.46&.38&.33&\textbf{.22}&\textbf{.11}&\textbf{.48}&\textbf{.38}&\textbf{.47}&\textbf{.37}&.14&.11\\
Tk-Instruct(3B)+SD&\textbf{.46}&.50&.15&\textbf{.89}&\textbf{.47}&\textbf{.52}&\textbf{.37}&.15&.08&.31&.32&.44&.30&\textbf{.43}&\textbf{.14}\\
Tk-Instruct(11B)&.35&\textbf{.14}&.09&.62&\textbf{.48}&.72&.56&\textbf{.36}&\textbf{.15}&.24&.19&.30&\textbf{.29}&.10&.09\\
Tk-Instruct(11B)+SD&\textbf{.37}&.13&\textbf{.10}&\textbf{.72}&.47&\textbf{.74}&\textbf{.57}&.27&.13&\textbf{.27}&\textbf{.21}&\textbf{.33}&.28&\textbf{.11}&\textbf{.10}\\
\midrule
OPT-IML(1.3B)&.53&.53&.19&.38&.35&.57&.50&\textbf{.78}&\textbf{.43}&.60&\textbf{.31}&\textbf{.54}&\textbf{.45}&.28&.24\\
OPT-IML(1.3B)+SD&\textbf{.54}&\textbf{.61}&\textbf{.21}&\textbf{.40}&\textbf{.37}&\textbf{.69}&\textbf{.59}&.69&.38&.60&.28&.47&.38&\textbf{.29}&.24\\
OPT-IML(30B)&.60&.50&.19&.58&.49&.80&.69&.84&.54&\textbf{.67}&\textbf{.53}&\textbf{.57}&\textbf{.50}&.27&.24\\
OPT-IML(30B)+SD&\textbf{.63}&\textbf{.55}&\textbf{.20}&\textbf{.64}&\textbf{.53}&\textbf{.85}&\textbf{.74}&\textbf{.86}&\textbf{.57}&.65&.51&.52&.39&\textbf{.35}&\textbf{.29}\\
\midrule
Dolly-V2(2.8B)&\textbf{.22}&\textbf{.33}&\textbf{.17}&.13&.13&.25&.25&.14&.06&\textbf{.25}&\textbf{.25}&\textbf{.36}&\textbf{.19}&.05&.06\\
Dolly-V2(2.8B)+SD&.19&.09&.06&\textbf{.14}&\textbf{.14}&.25&.25&.14&.06&.21&.20&.35&.18&\textbf{.12}&\textbf{.11}\\
Dolly-V2(6.9B)&\textbf{.26}&\textbf{.50}&\textbf{.17}&.14&.14&\textbf{.20}&\textbf{.20}&.14&.06&.30&.23&.31&.27&\textbf{.20}&\textbf{.20}\\
Dolly-V2(6.9B)+SD&.21&.30&.15&\textbf{.15}&\textbf{.15}&.16&.16&.14&.06&\textbf{.31}&.23&\textbf{.33}&\textbf{.28}&.07&.06\\
Dolly-V2(12B)&\textbf{.32}&.07&.06&.22&.22&\textbf{.47}&\textbf{.43}&\textbf{.38}&\textbf{.22}&.58&\textbf{.31}&\textbf{.23}&\textbf{.17}&\textbf{.27}&\textbf{.14}\\
Dolly-V2(12B)+SD&.28&\textbf{.08}&\textbf{.08}&\textbf{.27}&\textbf{.26}&.43&.41&.16&.07&\textbf{.59}&.26&.21&.15&.25&.12\\
\bottomrule
    \end{tabular}
    }
    \caption{Comparison of zero-shot prompting performance using hard-label evaluation with (SD) and without sociodemographic information. F1 is macro-averaged F1 and Acc is for Accuracy. Bold scores highlight the better performance when comparing the same model. Avg denotes averaged accuracy scores.}
    \label{app:tab:zeroshot_hard_all}
\end{table*}

\begin{table*}[]
    \centering
    \resizebox{1.00\textwidth}{!}{%
    \begin{tabular}{lc|cc|cc|cc|cc|cc|cc|cc|cc|cc|cc|cc|cc|cc|cc}
    \toprule
        & & \multicolumn{4}{c}{Toxicity} & \multicolumn{4}{c}{Hatespeech} & \multicolumn{4}{c}{Stance Detection} & \multicolumn{2}{c}{Sentiment}  \\
         & & \multicolumn{2}{c}{DP} & \multicolumn{2}{c}{Jigsaw} & \multicolumn{2}{c}{GHC} & \multicolumn{2}{c}{H-Twitter} & \multicolumn{2}{c}{SE2016} & \multicolumn{2}{c}{GWSD} & \multicolumn{2}{c}{Diaz} \\
         Model & Avg & CE & JSD & CE & JSD & CE & JSD & CE & JSD & CE & JSD & CE & JSD & CE & JSD \\
\midrule
InstructGPT(175B)& .21 &1.42&.32&.55&.15&.43&.07&\textbf{.88}&.08&\textbf{.98}&.27&\textbf{.86}&.18&1.50&.37\\
InstructGPT(175B)+SD& \textbf{.18} &\textbf{1.40}&\textbf{.29}&\textbf{.51}&\textbf{.12}&\textbf{.42}&\textbf{.06}&.90&.08&.99&\textbf{.25}&.89&\textbf{.15}&\textbf{1.48}&\textbf{.33}\\
\midrule
Flan-T5(80M)& .46&1.34&.26&1.06&.50&.98&.45&1.71&.68&1.25&.46&1.04&.29&1.75&.59\\
Flan-T5(80M)+SD&\textbf{.41} &1.34&\textbf{.25}&\textbf{.96}&\textbf{.41}&\textbf{.66}&\textbf{.20}&1.71&.68&\textbf{1.23}&\textbf{.43}&1.04&.29&1.75&\textbf{.58}\\
Flan-T5(250M)&.42 &1.65&.48&\textbf{.88}&\textbf{.37}&1.10&.53&1.14&.26&\textbf{1.00}&.29&\textbf{1.22}&.44&1.70&.54\\
Flan-T5(250M)+SD&\textbf{.40} &1.65&.48&1.14&.55&\textbf{1.06}&\textbf{.49}&\textbf{.94}&\textbf{.12}&1.01&\textbf{.28}&1.23&\textbf{.42}&\textbf{1.67}&\textbf{.49}\\
Flan-T5(780M)&.44 &\textbf{1.65}&.48&\textbf{.75}&\textbf{.29}&\textbf{.97}&.44&1.68&.66&\textbf{1.16}&.41&\textbf{1.06}&.30&1.65&.50\\
Flan-T5(780M)+SD&\textbf{.42} &1.67&\textbf{.45}&1.03&.44&1.00&.44&\textbf{1.64}&\textbf{.60}&1.25&.41&1.07&\textbf{.28}&\textbf{1.51}&\textbf{.35}\\
Flan-T5(3B)&.42 &\textbf{1.34}&\textbf{.26}&\textbf{.64}&\textbf{.22}&\textbf{.94}&\textbf{.42}&\textbf{1.63}&\textbf{.61}&\textbf{1.25}&.48&1.18&.42&1.72&.56\\
Flan-T5(3B)+SD&\textbf{.40} &1.54&.37&.75&.25&1.05&.47&1.70&.67&1.28&\textbf{.47}&\textbf{1.06}&\textbf{.26}&\textbf{1.49}&\textbf{.34}\\
Flan-T5(11B)&\textbf{.38} &\textbf{1.60}&.45&\textbf{.86}&\textbf{.36}&\textbf{.89}&\textbf{.39}&\textbf{1.10}&\textbf{.24}&1.21&.44&1.11&.34&\textbf{1.55}&.41\\
Flan-T5(11B)+SD&.39 &1.66&\textbf{.43}&1.11&.49&1.08&.46&1.47&.43&\textbf{1.18}&\textbf{.34}&\textbf{1.10}&\textbf{.23}&1.59&\textbf{.35}\\
\midrule
Flan-UL(20B)& .45 & 1.49 & .37 & .67 & .24 & \textbf{.87} & \textbf{.38} & 1.71 & .67 & 1.32 & .51 & 1.13 & .36 & 1.77 & .61\\
Flan-UL(20B)+SD& \textbf{.41} & \textbf{1.44} & .31 & .67 & \textbf{.21} & 1.03 & .46 & \textbf{1.69}& \textbf{.64} & \textbf{1.30} & \textbf{.48} & \textbf{1.12} & \textbf{.31} & \textbf{1.69} & \textbf{.49} \\
\midrule
Tk-Instruct(80M)& \textbf{.29}&\textbf{1.67}&\textbf{.50}&\textbf{.41}&.06&.42&.06&\textbf{1.07}&\textbf{.21}&\textbf{.98}&.28&\textbf{1.05}&.30&1.80&.64\\
Tk-Instruct(80M)+SD&.32 &1.73&.54&.42&.06&.42&.06&1.34&.37&1.00&.28&1.07&\textbf{.27}&1.80&.64\\
Tk-Instruct(250M)&\textbf{.33} &1.67&.52&\textbf{.40}&.05&\textbf{.43}&\textbf{.07}&\textbf{1.11}&\textbf{.24}&\textbf{1.23}&\textbf{.44}&1.11&.35&1.80&.64\\
Tk-Instruct(250M)+SD&.34 &\textbf{1.63}&\textbf{.45}&.41&.05&.47&.09&1.31&.35&1.25&.45&1.11&.35&1.80&.64\\
Tk-Instruct(780M)&.33 &\textbf{1.43}&\textbf{.32}&.58&.17&.56&.16&\textbf{1.00}&.17&\textbf{1.28}&.50&1.07&.32&1.80&.64\\
Tk-Instruct(780M)+SD&\textbf{.31} &1.66&.45&\textbf{.54}&\textbf{.12}&\textbf{.49}&\textbf{.09}&1.03&\textbf{.16}&1.29&\textbf{.44}&\textbf{1.06}&\textbf{.29}&\textbf{1.79}&\textbf{.63}\\
Tk-Instruct(3B)&.35& \textbf{1.36}&\textbf{.27}&.48&.10&.90&.40&\textbf{1.50}&.53&\textbf{1.07}&.35&\textbf{1.02}&.28&1.69&.54\\
Tk-Instruct(3B)+SD&\textbf{.29} &1.44&.29&\textbf{.43}&\textbf{.06}&\textbf{.77}&\textbf{.26}&1.56&\textbf{.52}&1.19&.35&1.05&\textbf{.25}&\textbf{1.47}&\textbf{.32}\\
Tk-Instruct(11B)&.4 &\textbf{1.67}&.50&.68&.24&.56&.16&\textbf{1.37}&\textbf{.43}&1.27&.49&1.19&.43&1.74&.58\\
Tk-Instruct(11B)+SD&\textbf{.36} &1.70&\textbf{.48}&\textbf{.58}&\textbf{.13}&.56&\textbf{.12}&1.46&.46&\textbf{1.25}&\textbf{.45}&\textbf{1.15}&\textbf{.34}&\textbf{1.71}&\textbf{.52}\\
\midrule
OPT-IML(1.3B)&.29 &1.40&.31&.90&.39&.72&.27&\textbf{.96}&\textbf{.13}&.97&.27&\textbf{.98}&.26&1.57&.43\\
OPT-IML(1.3B)+SD&\textbf{.25} &\textbf{1.36}&\textbf{.24}&\textbf{.88}&\textbf{.32}&\textbf{.63}&\textbf{.16}&1.07&.15&.97&\textbf{.26}&1.03&\textbf{.25}&1.57&\textbf{.36}\\
OPT-IML(30B)&.24 &1.43&.33&.71&.26&.52&.13&.90&.10&\textbf{.90}&.22&\textbf{.95}&.23&1.57&.42\\
OPT-IML(30B)+SD&\textbf{.20} &\textbf{1.40}&\textbf{.29}&\textbf{.66}&\textbf{.20}&\textbf{.48}&\textbf{.09}&\textbf{.89}&\textbf{.07}&.91&\textbf{.21}&.99&.23&\textbf{1.52}&\textbf{.32}\\
\midrule
Dolly-V2(2.8B)&.49 &\textbf{1.55}&\textbf{.43}&1.14&.55&1.03&.48&1.55&.55&\textbf{1.24}&.46&\textbf{1.10}&.35&1.77&.61\\
Dolly-V2(2.8B)+SD&\textbf{.48} &1.73&.54&\textbf{1.11}&\textbf{.52}&1.03&\textbf{.47}&1.55&.55&1.25&\textbf{.43}&1.11&.35&\textbf{1.69}&\textbf{.49}\\
Dolly-V2(6.9B)&.47 &\textbf{1.42}&\textbf{.32}&1.13&.55&\textbf{1.09}&.52&1.55&.54&1.22&.45&1.19&.42&\textbf{1.63}&\textbf{.48}\\
Dolly-V2(6.9B)+SD&\textbf{.44} &1.57&.35&\textbf{1.10}&\textbf{.49}&1.10&\textbf{.50}&1.55&.54&\textbf{1.20}&\textbf{.36}&\textbf{1.17}&\textbf{.33}&1.73&.53\\
Dolly-V2(12B)&.43 &1.76&.58&1.05&.50&\textbf{.83}&.34&\textbf{1.33}&\textbf{.40}&.98&.28&1.27&.49&\textbf{1.58}&.43\\
Dolly-V2(12B)+SD&\textbf{.40} &\textbf{1.73}&\textbf{.48}&\textbf{.97}&\textbf{.38}&.86&\textbf{.31}&1.51&.50&.98&\textbf{.26}&1.27&\textbf{.45}&1.59&\textbf{.39}\\
\bottomrule
    \end{tabular}
    }
    \caption{Comparison of zero-shot prompting performance using soft-label evaluation with (SD) and without sociodemographic information. CE is cross-entropy and JSD is for Jensen-Shannon divergence. Bold scores highlight the better performance when comparing the same model. Avg denotes averaged JSD scores.}
    \label{app:tab:zeroshot_soft_all}
\end{table*}

\subsection{Influence of model choice on prediction outcomes}
\label{app:sec:model_choice_influence}

\begin{figure*}[t!]
    \centering
    \includegraphics[clip, trim=0cm 0cm 0.8cm 0cm, width=1.0\textwidth]{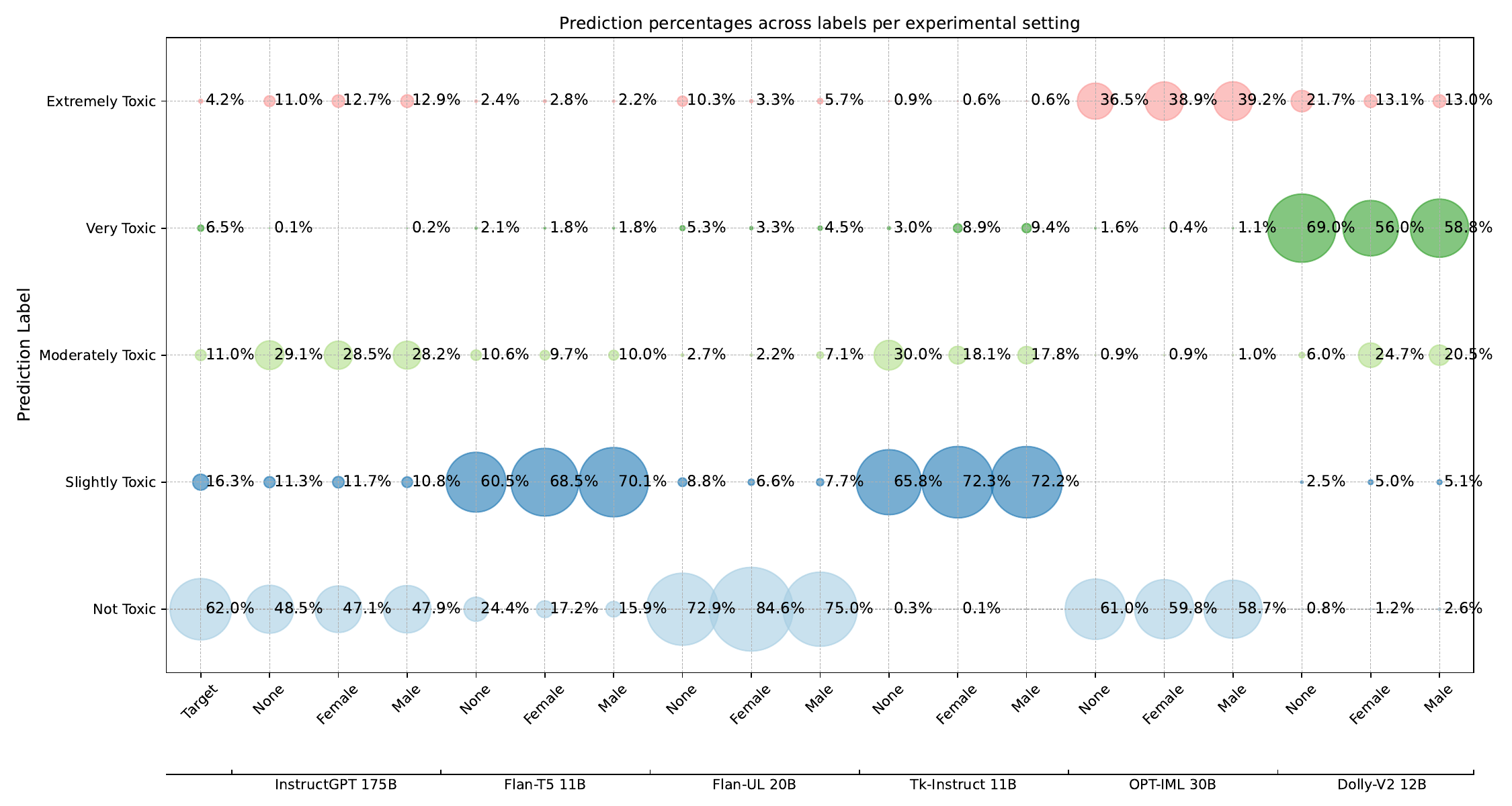}
    \caption{Prediction distributon across the different labels for the DP dataset. We compare the true label distribution (\texttt{Target}) with the results of different experimental settings for different models. \texttt{None} refers to prompting without sociodemographic information. \texttt{Female} and \texttt{Male} refer to sociodemographic prompting with a single attribute, respectively. The model choice has a larger influence on label predictions than the sociodemographic profile.}
    \label{app:fig:toxicity_label_space}
\end{figure*}

\begin{figure*}[t!]
    \centering
    \includegraphics[clip, trim=0cm 0cm 1cm 0cm, width=1.0\textwidth]{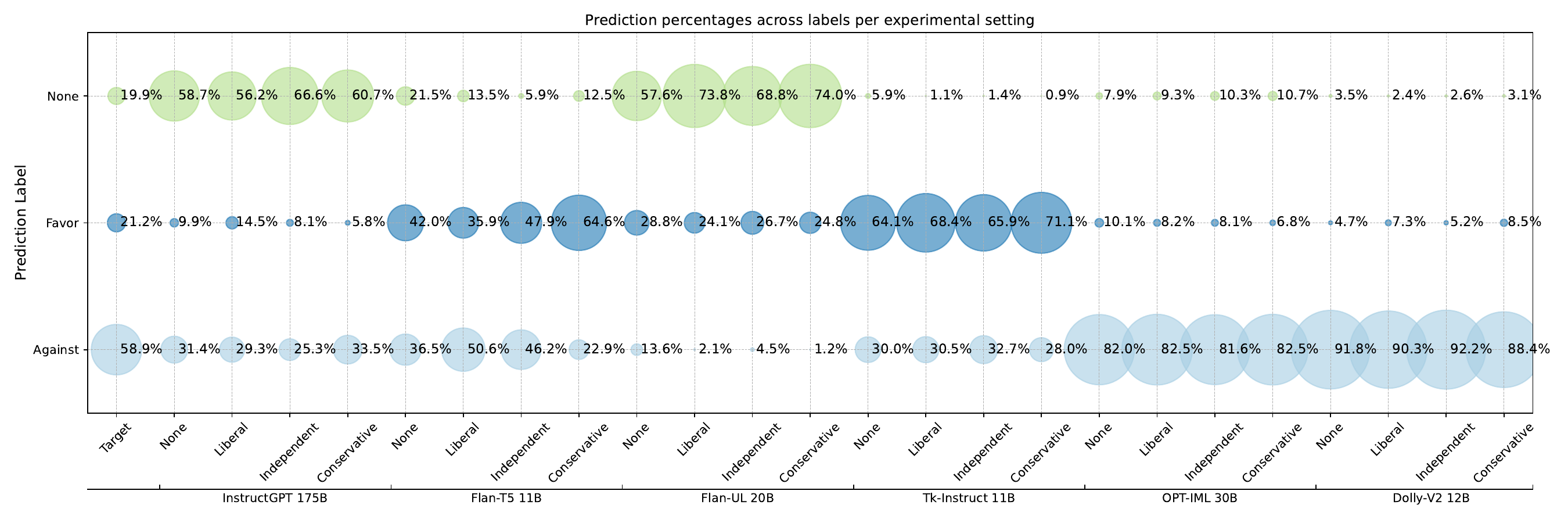}
    \caption{Prediction distributon across the different labels for the SE2016 dataset. We compare the true label distribution (\texttt{Target}) with the results of different experimental settings for different models. \texttt{None} refers to prompting without sociodemographic information. \texttt{Liberal}, \texttt{Independent} and \texttt{Conservative} refer to sociodemographic prompting with a single attribute, respectively. The model choice has a larger influence on label predictions than the sociodemographic profile.}
    \label{app:fig:stance_label_space}
\end{figure*}

We provide more detailed model comparisons of prediction outcomes in Figure~\ref{app:fig:toxicity_label_space} for the DP dataset and in Figure~\ref{app:fig:stance_label_space} for the SE2016 dataset, respectively.
For both datasets and the sociodemographic attributes tested (\texttt{Gender} for DP, \texttt{Political Affiliation} for SE2016), we observe that the model choice has a larger influence on the label prediction than the value of the sociodemographic attribute.

\subsection{Robustness Analysis}
\label{app:sec:robustness}

\begin{table*}[]
    \centering
    \resizebox{1.00\textwidth}{!}{%
    \begin{tabular}{c|p{20cm}}
    \toprule
       Identifier & Format  \\
       \midrule
       1 & Given a text, how would a person of gender \{gender\}, race '\{race\}', age '\{age\}', education level '\{education\}' and political affiliation '\{political\_affiliation\}' rate the degree of toxicity in the text. Possible values are 'not toxic', 'slightly toxic', 'moderately toxic', 'very toxic' or 'extremely toxic'. \newline Text: '\{text\}' \newline Toxicity: \\
       \midrule
       2 & Imagine you are a person of gender \{gender\}, race '\{race\}', age '\{age\}', education level '\{education\}' and political affiliation '\{political\_affiliation\}'. Given a text, how would you rate the degree of toxicity in the text. Possible values are 'not toxic', 'slightly toxic', 'moderately toxic', 'very toxic' or 'extremely toxic'. \newline Text: '\{text\}' \newline Toxicity: \\
       \midrule
       3 & Person: \{gender\}, \{race\}, \{age\}, \{education\}, \{political\_affiliation\} \newline Text: '\{text\}' \newline Toxicity ('not toxic'|'slightly toxic'|'moderately toxic'|'very toxic'|'extremely toxic'): \\
       \bottomrule
    \end{tabular}
    }
    \caption{Three different prompt formulations which are used to study the robustness of sociodemographic prompting. The placeholders in curly brackets are replaced with the respective values before prompting the LLM. The prompt formulations shown are exemplary for the DP dataset for toxicity detection.}
    \label{tab:robustness:templates}
\end{table*}

\begin{figure*}[t!]
    \centering
    \includegraphics[clip, trim=0cm 0cm 2cm 0cm, width=1.0\textwidth]{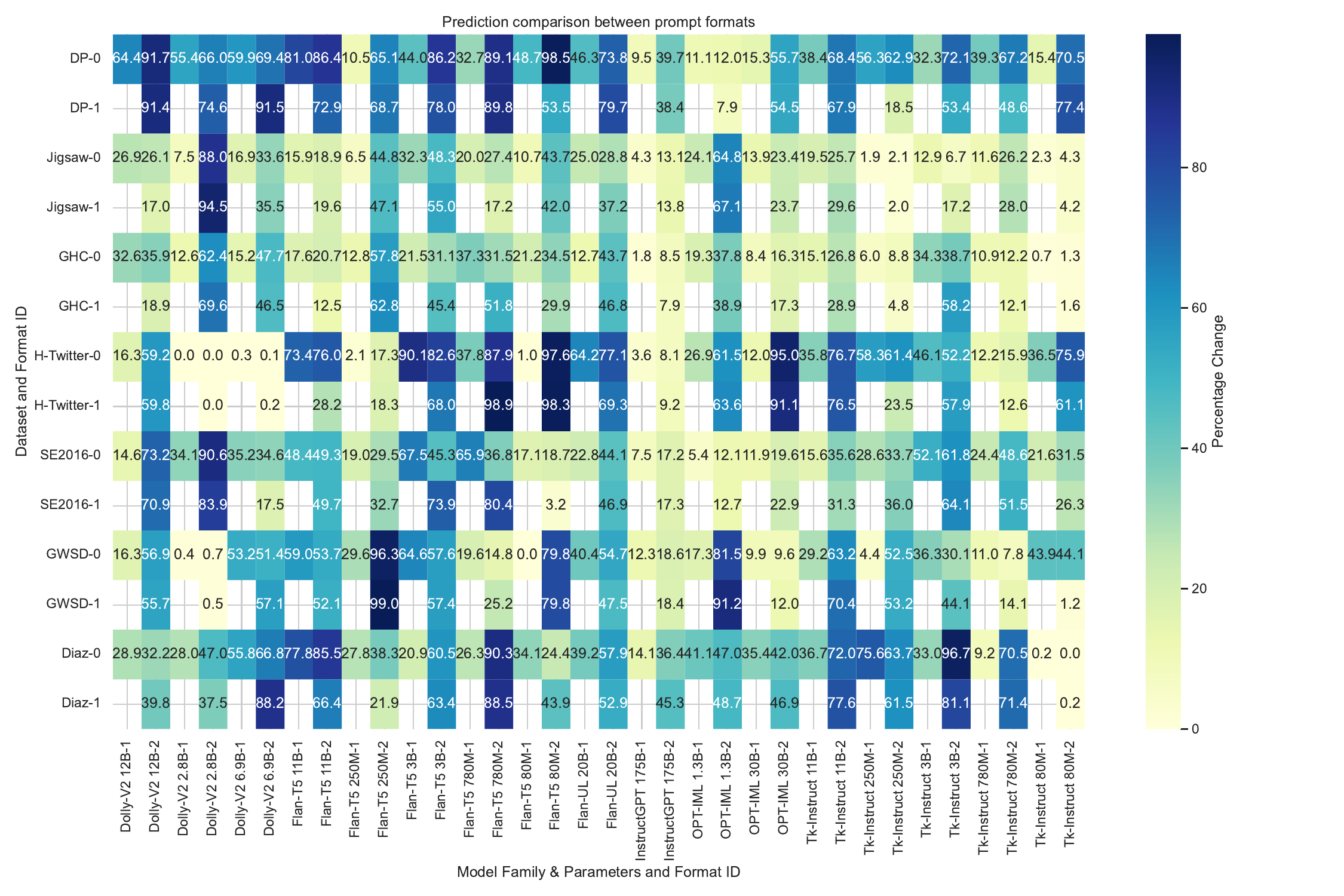}
    \caption{Comparison of label changes when prompting (with sociodemographic information) with different prompt formulations (i.e. format 0,1 or 2) across seven datasets and two models. A cell value resembles the prediction difference between prompting with the format id provided per row and column.}
    \label{app:fig:robustness_with_sd}
\end{figure*}

\begin{figure*}[t!]
    \centering
    \includegraphics[width=1.0\textwidth]{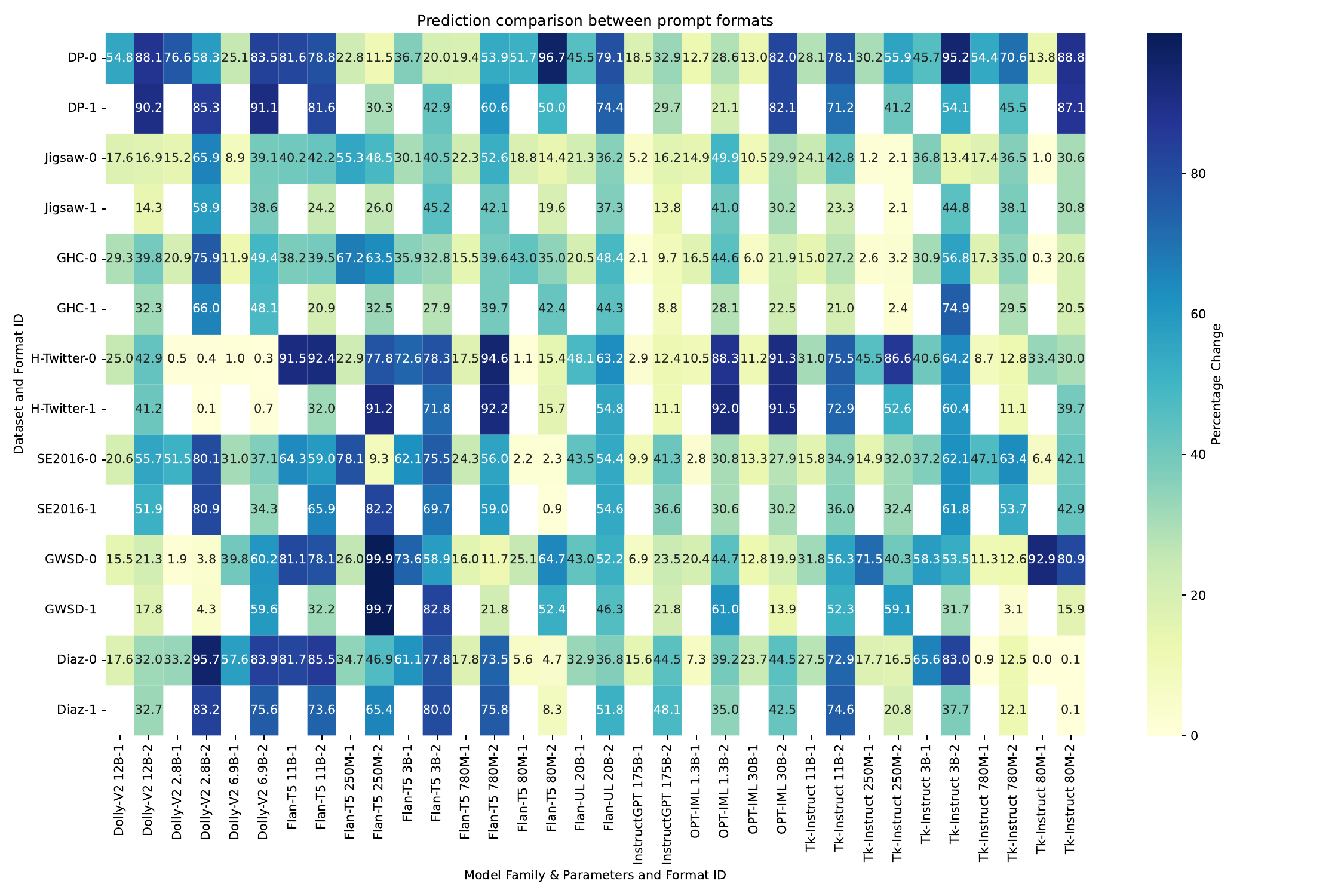}
    \caption{Comparison of label changes when prompting (without sociodemographic) with information different prompt formulations (i.e. format 0,1 or 2) across seven datasets and two models. A cell value resembles the prediction difference between prompting with the format id provided per row and column.}
    \label{app:fig:robustness_without_sd}
\end{figure*}

Several works demonstrated the brittleness of zero-shot and few-shot predictions of language models~\citep{min-etal-2022-rethinking} due to factors such as the prompt format or the order of the in-context examples~\citep{pmlr-v139-zhao21c, lu-etal-2022-fantastically}.
Thus, we evaluate if our results are subject to such variations, by repeating our experiments using three different prompt formulations.
The prompt formulations are displayed in Table~\ref{tab:robustness:templates}.
The complete results are shown in Figure~\ref{app:fig:robustness_with_sd} (with sociodemographic information) and Figure~\ref{app:fig:robustness_without_sd} (without sociodemographic information).

We observe that most differences are induced when comparing the prompt using a minimal formulation (format 2) with the more elaborate versions (format 0,1).
Structurally, we see that prompting both with and without sociodemographic information is affected by the prompt formulation to a large extent.

\hs{

\subsection{GLMM Analyses}\label{sec:glmm_analyses}
In addition to the reported percentages of label changes (\Cref{fig:label_changes}), classification performance measures (\Cref{tab:zeroshot_original_annotations,tab:zeroshot_performance}), and disagreement prediction performance (\Cref{fig:dispersion_heatmap}), we conduct statistical analyses using generalized linear mixed models (GLMMs).
GLMMs allow us to statistically account for various fixed and random effects and, thereby, account for potential confounders and statistical dependencies in our data.
We thus fit four GLMs/GLMMs for (i) the model sensitivity experiment (\Cref{ssec:model_sensitivity}), (ii) the prediction of individual (original) annotations (\Cref{ssec:zeroshot}), (iii) the prediction of aggregated annotations (\Cref{ssec:zeroshot}), and (iv) the prediction of ambiguous instances (\Cref{app:sec:ambig_instances}).
We model the respective binary label changes, classification accuracies, and disagreement prediction accuracies on an instance level across datasets.
We use R \citep{R} and mgcv 1.8-42 \citep{mgcv1,mgcv2,mgcv3,wood2017generalized,mgcv5} to fit all our models.
\subsubsection{Model Specifications}
We fit binomial models (logit link) and include the factors
(a) model family (e.g., Flan-T5),
(b) model size (logarithmic),
(c) task (e.g., sentiment analysis),
(d) text length in characters (i.e., length of the text to be classified), and
(e) additional prompt length in characters (i.e., length of the entire prompt excluding the text to be classified) as fixed effects for all models.
We additionally model the specific dataset as a random effect to account for structural dependencies stemming from the choice of dataset for all models except the (second) model that predicts individual annotations as, for this experiment, there only is one dataset per task.
In the following, we specify the predictor variable and additional covariates for the four models. 
\paragraph{Sensitivity Model.}
We model the probability of a label change between standard and SD prompting.
\paragraph{Individual Annotations Model.}
We model the probability of a model predicting the correct (individually-annotated) class label.
We additionally include the prompting method used (standard or sociodemographic prompting) as a fixed effect.
To gain further insights on the interaction of using sociodemographic prompting with model size, family, or text length, we additionally model all pairwise interactions between prompting method and the fixed effects listed above.
\paragraph{Aggregated Annotations Model.}
In this model, we predict the probability of a model predicting the correct aggregated label.
We include the same model terms as the previous model and additionally include a random effect of dataset.
\paragraph{Ambiguity Model.}
We model the probability of successfully predicting annotator disagreement as the predictor variable.
The included fixed and random effects are identical to the sensitivity model.
\subsubsection{Results}
\begin{table}[ht]
\centering
\resizebox{\columnwidth}{!}{
\begin{tabular}{rrrr}
  \toprule
Term & df & $\chi^2$ & p\\ 
  \midrule
  model size & 1 & 2769.41 & \textbf{<0.001} \\ 
  text length & 1 & 7.15 & \textbf{0.01} \\ 
  additional prompt length & 1 & 2.35 & 0.13 \\ 
  model family & 5 & 3937.76 & \textbf{<0.001} \\ 
  task & 3 & 2.72 & 0.44 \\ 
  \bottomrule
\end{tabular}
}\caption{Test statistics for the parametric fixed terms of the sensitivity model.}\label{tab:sensitivity_model_wald}
\end{table}
\begin{table}[ht]
\centering
\resizebox{\columnwidth}{!}{
\begin{tabular}{rrrr}
  \toprule
Term & df & $\chi^2$ & p\\ 
  \midrule
  model size & 1 & 564.29 & \textbf{<0.001} \\ 
  text length & 1 & 28.88 & \textbf{<0.001} \\ 
  additional prompt length & 1 & 0.70 & 0.40 \\ 
  model family & 5 & 579.84 & \textbf{<0.001} \\ 
  task & 3 & 5.92 & 0.12 \\  
  \bottomrule
\end{tabular}
}\caption{Test statistics for the parametric fixed terms of the ambiguity prediction model.}\label{tab:ambiguity_wald}
\end{table}
\paragraph{Sensitivity Model.}
\Cref{tab:sensitivity_model_wald} displays the test statistics regarding the parametric fixed terms of the sensitivity model.
We observe
a statistically significant positive effect of \textbf{model size} ($\beta$=0.56, 95\% CI [0.54, 0.58], $p$<0.001),
a significant negative effect of \textbf{text length} ($\beta$=-1.69e-04, 95\% CI [-2.93e-04, -4.51e-05], $p$=0.007), and
a statistically significant effect of \textbf{model family} ($\chi^2(5)$=3937.76, $p$<0.001).
A post hoc Wald comparison of the contrasts for model family revealed significant differences between all pairs of model families.
The corresponding estimates are (in descending order)
Flan-T5 ($\beta$=0.58, 95\% CI [0.53, 0.63]),
Tk-Instruct ($\beta$=0.33, 95\% CI [0.28, 0.37]),
Flan-UL ($\beta$=-0.21, 95\% CI [-0.27, -0.14]),
OPT-IML ($\beta$=-0.64, 95\% CI [-0.70, -0.59]), and
InstructGPT ($\beta$=-1.90, 95\% CI [-1.99, -1.82]).
Note that the estimate for Dolly-V2 is fixed as the reference level.
\paragraph{Individual Annotations Model.}
\Cref{tab:individual_annotation_wald} displays the test statistics regarding the parametric fixed terms of the individual annotation prediction model.
As, in contrast to the sensitivity model, that modelled the effect of SD prompting within the predictor variable, this model includes SD prompting as a covariate, we focus on effects involving the prompting method variable.\footnote{The interpretation of, e.g., the statistically significant effect of text length is that it has an overall impact on prediction accuracy for both, standard prompting as well as SD prompting. Instead of this main effect, we are interested in the interaction effect, i.e., the relevance of text length particularly when SD prompting is used.}
While we do not find a significant main effect of using SD prompting, we observe several interaction effects of SD prompting.
Concretely, we observe
a statistically significant negative interaction effect of \textbf{model size and SD prompting} ($\beta$=-0.12, 95\% CI [-0.15, -0.09], $p$<0.001),
a statistically significant positive interaction effect of \textbf{text length and SD prompting} ($\beta$=6.95e-04, 95\% CI [4.91e-04, 8.99e-04], $p$<0.001), and
a statistically significant negative interaction effect of \textbf{model family and SD prompting} ($\chi^2(5)$=561.33, $p$<0.001).
A post hoc Wald comparison of the contrasts for model family using SD prompting revealed significant differences between all pairs of model families.
The corresponding estimates are (in descending order)
\instructgpt ($\beta$=-0.02, 95\% CI [-0.13, 0.10]),
\imlopt ($\beta$=-0.13, 95\% CI [-0.23, -0.03]),
\flant ($\beta$=-0.63, 95\% CI [-0.73, -0.53]),
\tkinstruct ($\beta$=-0.73, 95\% CI [-0.84, -0.63]) and
\dolly ($\beta$=-0.85, 95\% CI [-0.95, -0.75]).
Note that the estimate for \flantul is fixed as the reference level.
\paragraph{Aggregated Annotations Model.}
\Cref{tab:aggregated_annotation_wald} displays the test statistics regarding the parametric fixed terms of the aggregated annotation prediction model.
As for the previous model, we focus our discussion on effects involving SD prompting.
We observe
a statistically significant negative interaction effect of \textbf{model size and SD prompting} ($\beta$=-0.08, 95\% CI [-0.10, -0.05], $p$<0.001),
a statistically significant negative interaction effect of \textbf{additional prompt length and SD prompting} ($\beta$=-1.71e-03, 95\% CI [-2.36e-03, -1.06e-03], $p$<0.001),
a statistically significant interaction effect of \textbf{model family and SD prompting} ($\chi^2(5)$=101.29, $p$<0.001), and
a statistically significant interaction effect of \textbf{task and SD prompting} ($\chi^2(3)$=309.46, $p$<0.001).
For model family, a post hoc Wald comparison of the contrasts for model family using SD prompting revealed significant differences between all pairs of model families except \flantul and \flant.
The corresponding estimates are (in descending order)
\instructgpt ($\beta$=0.06, 95\% CI [-0.05, 0.17]),
\imlopt ($\beta$=0.01, 95\% CI [-0.08, 0.10]),
\tkinstruct ($\beta$=-0.20, 95\% CI [-0.28, -0.11]).
\dolly ($\beta$=-0.22, 95\% CI [-0.31, -0.13]), and
\flant ($\beta$=-0.26, 95\% CI [-0.35, -0.17]),
Note that the estimate for \flantul is fixed as the reference level.
For task, a post hoc Wald comparison of the contrasts for task using SD prompting revealed significant differences between hatespeech and sentiment, sentiment and toxicity, and sentiment and stance.
The corresponding estimates are (in descending order)
sentiment ($\beta$=0.58, 95\% CI [0.50, 0.65]),
stance ($\beta$=0.17, 95\% CI [0.11, 0.22]),
toxicity ($\beta$=-0.06, 95\% CI [-0.10, -8.25e-03]), and
Note that the estimate for hatespeech is fixed as the reference level.
\paragraph{Ambiguity Model.}
\Cref{tab:ambiguity_wald} displays the test statistics regarding the parametric fixed terms of the disagreement prediction model.
We observe
a statistically significant positive effect of \textbf{model size} ($\beta$=0.22, 95\% CI [0.21, 0.24], $p$<0.001),
a significant negative effect of \textbf{text length} ($\beta$=-2.72e-04, 95\% CI [-3.71e-04, -1.73e-04], $p$<0.001), and
a statistically significant effect of \textbf{model family} ($\chi^2(5)$=579.84, $p$<0.001).
A post hoc Wald comparison of the contrasts for model family revealed significant differences between all pairs of model families except \flantul and \instructgpt.
The corresponding estimates are (in descending order)
\dolly ($\beta$=0.58, 95\% CI [0.52, 0.64]),
\flant ($\beta$=0.53, 95\% CI [0.47, 0.58]),
\tkinstruct ($\beta$=0.40, 95\% CI [0.34, 0.45]),
\imlopt ($\beta$=0.30, 95\% CI [0.24, 0.36]), and
\instructgpt ($\beta$=0.03, 95\% CI [-0.04, 0.10]),
Note that the estimate for \flantul is fixed as the reference level.
\begin{table*}[h]
\centering
\resizebox{0.65\textwidth}{!}{
\begin{tabular}{rrrr}
  \toprule
Term & df & $\chi^2$ & p\\ 
  \midrule
  model size & 1 & 0.44 & 0.51 \\ 
  text length & 1 & 10.98 & \textbf{<0.001} \\ 
  additional prompt length & 1.00 & 0.01 & 0.92 \\ 
  model family & 5 & 990.68 & \textbf{<0.001} \\ 
  task & 1 & 1.31 & 0.25 \\ 
  prompting method & 1 &  0.0 & 1.0 \\ 
  model size : prompting method & 1 & 50.61 & \textbf{<0.001} \\ 
  text length : prompting method & 1 & 44.54 & \textbf{<0.001} \\ 
  additional prompt length : prompting method & 1 & 0.01 & 0.93 \\ 
  model family : prompting method & 5 & 561.33 & \textbf{<0.001} \\ 
  task : prompting method & 1 & 0.00 & 0.99 \\
  \bottomrule
\end{tabular}
}\caption{Test statistics for the parametric fixed terms of the individual annotation prediction model.}\label{tab:individual_annotation_wald}
\end{table*}
\begin{table*}[ht]
\centering
\resizebox{0.65\textwidth}{!}{
\begin{tabular}{rrrr}
  \toprule
Term & df & $\chi^2$ & p\\ 
  \midrule
  model size & 1 & 3.09 & 0.08 \\ 
  text length & 1 & 6.42 & \textbf{0.01} \\ 
  additional prompt length & 1 & 0.83 & 0.36 \\ 
  model family & 5 & 6470.73 & \textbf{<0.001} \\ 
  task & 3 & 18.31 & \textbf{<0.001} \\ 
  prompting method & 1 & 45.40 & \textbf{<0.001} \\ 
  model size : prompting method & 1 & 30.04 & \textbf{<0.001} \\ 
  text length : prompting method & 1 & 2.79 & 0.09 \\ 
  additional prompt length : prompting method & 1 & 26.84 & \textbf{<0.001} \\ 
  model family : prompting method & 5 & 101.29 & \textbf{<0.001} \\ 
  task : prompting method & 3 & 309.46 & \textbf{<0.001} \\ 
  \bottomrule
\end{tabular}
}\caption{Test statistics for the parametric fixed terms of the aggregated annotation prediction model.}\label{tab:aggregated_annotation_wald}
\end{table*}
}

\end{document}